\documentclass[journal, 10pt, twoside]{IEEEtran}
\usepackage{amsmath,amsfonts}
\usepackage{url}
\usepackage{graphicx}
\usepackage{cite}
\usepackage{booktabs}
\usepackage{multirow}
\usepackage{pifont}
\usepackage{setspace}
\usepackage{enumitem}
\usepackage[switch]{lineno}
\usepackage{float}
\usepackage{ulem}
\makeatletter
\let\MYcaption\@makecaption
\usepackage[font=footnotesize]{subcaption}
\let\@makecaption\MYcaption
\usepackage{hyperref}
\hypersetup{
    colorlinks=true,
    linkcolor=blue,
    filecolor=black,
    urlcolor=black,
    citecolor=purple
}
\usepackage[ruled,noend]{algorithm2e}
\SetAlCapNameFnt{\small}
\SetAlCapFnt{\small}
\usepackage[table,xcdraw]{xcolor}
\newcommand{\egi}{\textit{e.g.}}
\begin{document}

\title{These Maps Are Made by Propagation:\\Adapting Deep Stereo Networks to Road Scenarios with Decisive Disparity Diffusion}

\normalem
\author{Chuang-Wei Liu, Yikang Zhang, Qijun Chen,~\IEEEmembership{Senior Member,~IEEE},\\ Ioannis Pitas,~\IEEEmembership{Life Fellow,~IEEE}, and Rui Fan,~\IEEEmembership{Senior Member,~IEEE}
}

\maketitle

\begin{abstract}
Stereo matching has emerged as a cost-effective solution for road surface 3D reconstruction, garnering significant attention towards improving both computational efficiency and accuracy. {This article introduces decisive disparity diffusion (D3Stereo), marking the first exploration of dense deep feature matching that adapts pre-trained deep convolutional neural networks (DCNNs) to previously unseen road scenarios}. A pyramid of cost volumes is initially created using various levels of learned representations. Subsequently, a novel recursive bilateral filtering algorithm is employed to aggregate these costs. A key innovation of D3Stereo lies in its alternating decisive disparity diffusion strategy, wherein intra-scale diffusion is employed to complete sparse disparity images, while inter-scale inheritance provides valuable prior information for higher resolutions. Extensive experiments conducted on our created UDTIRI-Stereo and Stereo-Road datasets underscore the effectiveness of D3Stereo strategy in adapting pre-trained DCNNs and its superior performance compared to all other explicit programming-based algorithms designed specifically for road surface 3D reconstruction. Additional experiments conducted on the Middlebury dataset with backbone DCNNs pre-trained on the ImageNet database further validate the versatility of D3Stereo strategy in tackling general stereo matching problems.
\end{abstract}


\section{Introduction}
\label{sec.intro}

\IEEEPARstart{E}{nsuring} safe and comfortable driving requires the timely assessment of road conditions and the prompt repair of road defects \cite{ma2022computer}. With an increasing emphasis on maintaining high-quality road conditions \cite{liang2022automatic}, the demand for automated 3D road data acquisition systems has grown more intense than ever \cite{fan2019road,haq2019stereo}. The study presented in \cite{ahmed2021pothole} employs a laser scanner to collect high-precision 3D road data. Nevertheless, the high equipment costs and the long-term maintenance expenses have limited the widespread adoption of such laser scanner-based systems \cite{fan2019pothole}. {Therefore, stereo vision, a process similar to human binocular vision that provides depth perception using dual cameras, has emerged as a practical and cost-effective alternative for accurate 3D road data acquisition \cite{fan2018road,fan2021graph}.} Existing stereo matching approaches are either explicit programming-based or data-driven. The former ones rely on hand-crafted feature extraction and estimate disparities through local block matching or global energy minimization \cite{liu2024playing}. Nonetheless, hand-crafted feature extraction faces challenges in handling varying lighting conditions and noises. With recent advances in deep learning, researchers have resorted to deep convolutional neural networks (DCNNs) for stereo matching \cite{PSMNet,RAFT}. These data-driven approaches can learn abstract features directly from input stereo images, making them increasingly favored in this research domain. Unfortunately, the limited availability of well-annotated road disparity data restrains the transfer learning of these DCNNs \cite{liu2023stereo}. Therefore, explicitly programming-based stereo matching approaches \cite{fan2018road,FBS,fan2021rethinking} remain the mainstream in the field of road surface 3D reconstruction.

Building upon the local coherence constraint \cite{roy1999stereo}, seed-and-grow stereo matching algorithms \cite{miksik2015incremental, pillai2016high, fan2018road} have been widely utilized for quasi-dense disparity estimation. Given that road disparities change gradually across continuous regions, our previously published road surface 3D reconstruction algorithm search range propagation (SRP) \cite{fan2018road} initializes disparity seeds using a winner-take-all (WTA) strategy at the bottom row of the image and estimates disparities iteratively with the search range propagated from three neighboring seeds. Another significant contribution of \cite{fan2018road} lies in the perspective transformation (PT), designed to convert the target view of road image into a reference view. {This transformation helps decrease computations by reducing the disparity search range and improving stereo matching accuracy by increasing the similarity of the compared blocks.} While the combination of SRP and PT yields a remarkable 3D geometry reconstruction accuracy of approximately 3 mm, it is noteworthy that the disparity estimation accuracy remains constrained by the reliability of the initial seeds generated using the simple WTA strategy. The unidirectional disparity propagation process further leads to disparity estimation errors on discontinuities, such as road defects. Additionally, both seed-and-grow stereo matching and perspective transformation require a set of sparse yet reliable initial correspondences, and the density and reliability of these correspondences directly affect the efficiency and accuracy of the seed-growing process.

Drawing inspiration from recent advances in plug-and-play sparse correspondence matching \cite{efe2021dfm,zhou2023e3cm} approaches, we propose a feasible solution to address these limitations. For example, the deep feature matching (DFM) method \cite{efe2021dfm} utilizes a backbone DCNN pre-trained on the ImageNet database \cite{russakovsky2015imagenet} to extract feature pyramids for both views, and subsequently refines the coarse correspondences initialized at the deepest feature layer to former layers following a linear hierarchical manner. These methods have demonstrated the effectiveness of using deep features provided by pre-trained backbones to solve the correspondence matching task. Therefore, our primary motivation is to develop a dense deep feature matching strategy by improving the seed-and-grow stereo matching with the hierarchical refinement strategy in DFM. Leveraging accurate sparse correspondences as disparity seeds, such a dense deep feature matching strategy exhibits compatibility with perspective transformation, thus leading to improvements in both stereo matching accuracy and efficiency compared with the combination of SRP and PT. However, directly incorporating a hierarchical refinement strategy into seed-and-grow stereo matching still has the following limitations:
\begin{itemize}
    \item The dense matching process in the stereo matching task requires additional matching noise elimination operations in challenging areas with weak/repetitive textures.
    \item Additional efforts in eliminating inaccurate sparse correspondences are required to mitigate error accumulation and propagation in the seed-growing process.
    \item The linear hierarchical refinement strategy in DFM is designed to enhance the spatial details of the coarse initial correspondences, while having limited effectiveness in enhancing their density. 
\end{itemize}

To address these limitations, we propose a plug-and-play stereo matching strategy for road surface 3D reconstruction, referred to as \uline{\textbf{D}ecisive \textbf{D}isparity \textbf{D}iffusion \textbf{Stereo} (\textbf{D3Stereo})}, serving as the first exploration of dense deep feature matching. D3Stereo is compatible with any hand-crafted feature extraction approaches, stereo matching networks pre-trained on other public datasets, and even backbone DCNNs pre-trained for image classification. We first propose the recursive bilateral filtering (RBF) algorithm, a more efficient alternative to traditional bilateral filtering (BF) \cite{FBS} for matching cost aggregation. By recursively applying a small filtering kernel, our BRF achieves a significantly expanded receptive field while maintaining the same computational cost as BF, thereby gathering more context information for cost aggregation. The proposed method leverages the powerful semantic feature extraction ability of a pre-trained DCNN backbone in a hierarchical manner. It consists of two algorithms that diffuse decisive disparities at both intra and inter scales, respectively. With a cost volume pyramid built with different layers of feature maps, we first find coarse decisive disparities at the deepest layer. Then, the coarse decisive disparities are adversarially diffused to their neighboring pixels in the same layer to yield a dense disparity map, within which reliable decisive disparities are inherited into the former layer by checking the matching cost local minima consistency between consecutive layers. Our adversarial disparity diffusion process and novel disparity inheritance strategy help in eliminating the inaccurate correspondences initialized at the last layer. Afterwards, the derived refinement results activate the decisive disparity intra and inter scale diffusion in the former layer. This process is repeated until a dense disparity map is obtained at the finest resolution layer. In general, the combined usage of diffusing decisive disparities at both intra and inter scales fully exploits the semantic information at different scales of feature maps and thus obtaining improved disparity seeds in terms of both accuracy and distribution uniformity compared with the hierarchical refinement strategy in DFM \cite{efe2021dfm}. 

Additionally, we create a synthetic road dataset called the UDTIRI-Stereo dataset using the CARLA simulator \cite{dosovitskiy2017carla} for disparity estimation evaluation. Although collecting datasets using simulators has emerged as a prevalent alternative for real-world datasets \cite{cabon2020virtual,Cre}, these simulators model the road surface as a ground plane, thereby significantly reducing the complexity of the stereo matching task. In order to narrow the domain gap between the idealized road surface in CARLA and the real-world road surface, we originally augment the road surface mesh model in CARLA with 1) 2D Perlin noise and 2) digital twins of pothole models. By applying linear interpolation between initial random noises, 2D Perlin noise is utilized to generate the natural undulations of the real-world road surface. Moreover, digital twins of pothole models yielded in real-world \cite{fan2019pothole} are randomly transplanted onto the road surface, thus further introducing disparity discontinuities into the UDTIRI-Stereo dataset. 

\section{Related Works}
\label{sec.related}

\subsection{Stereo Matching for Road Surface 3D Reconstruction}

Several stereo matching approaches \cite{fan2018road,FBS,fan2021rethinking}, developed specifically for road surface 3D reconstruction, have been proposed since 2014 \cite{zhang2014efficient}. The first reported effort in this area of research was our proposed iterative stereo matching algorithm SRP \cite{fan2018road}. Despite the remarkable 3D geometry reconstruction accuracy yielded by SRP, its row-by-row disparity propagation process is challenging to implement in parallel on GPUs. To address this issue, \cite{FBS} proposed a GPU-friendly algorithm for road disparity estimation based on fast bilateral stereo (FBS) that can be embedded in a drone for real-time road surface 3D reconstruction. Nevertheless, the bilateral filtering process in FBS is computationally intensive, especially when a large filter kernel is employed, thus further increasing the memory burden on the embedded computers. As a result, semi-global matching (SGM) was used in conjunction with PT in \cite{fan2021rethinking} for road disparity estimation. Experimental results suggest that SGM outperforms both SRP and FBS when PT is incorporated. In general, D3Stereo continues the search range propagation strategy in SRP, while the seed-growing process is executed within a single instruction multiple data architecture for better leveraging parallel computing resources. Additionally, a recursive bilateral filter is proposed for more efficient cost aggregation compared with FBS. 

On the other hand, recent domain generalization-aimed stereo matching networks \cite{rao2023masked,chang2023domain,zhang2022revisiting,liu2022graftnet,shen2021cfnet} achieve remarkable generalizability across various scenarios. Common strategies to maintain the performance of stereo matching networks under scene changes include narrowing the cross-domain feature inconsistency \cite{chang2023domain,liu2022graftnet,shen2021cfnet} and enhancing the ability of DCNNs to learn more image structure information \cite{rao2023masked}. However, stereo matching in road scenes additionally emphasizes the ability of networks to handle fine-grained disparity variations compared to general domain adaptation tasks, and the performance of these domain generalization-aimed stereo matching networks in road scenes remains unverified.

\subsection{Sparse Correspondence Matching}

Conventional hand-crafted sparse correspondence matching approaches \cite{lowe2004distinctive,bay2006surf,leutenegger2011brisk} first extract keypoints using explicitly designed local feature detectors and descriptors. Correspondence pairs are then determined using the nearest neighbor search algorithm. Although recent data-driven approaches \cite{barroso2022key, detone2018superpoint, dusmanu2019d2, revaud2019r2d2, sarlin2020superglue, rocco2020ncnet, rocco2020efficient, sun2021loftr} have demonstrated significant improvements over hand-crafted methods, these supervised methods usually demand a large amount of well-annotated data for model training, resulting in unsatisfactory performance when applied to new domains \cite{zhou2023e3cm}. Moreover, adopting an independent sparse correspondence matching algorithm, regardless of whether it relies on explicit programming or DCNNs, for seed initialization leads to increased consumption of memory and computational resources. Recent plug-and-play approaches, such as DFM \cite{efe2021dfm} and epipolar-constrained cascade correspondence matching (E3CM) \cite{zhou2023e3cm}, utilize backbone DCNNs pre-trained on the ImageNet database \cite{russakovsky2015imagenet} for sparse correspondence matching based on a hierarchical refinement strategy, obviating the necessity for model training and fine-tuning. Consequently, a preferred solution would be to opt for plug-and-play algorithms that leverage the backbone DCNN incorporated into D3Stereo for seed initialization.

\begin{figure*}[t!]
	\begin{center}
		\centering
		\includegraphics[width=1\textwidth]{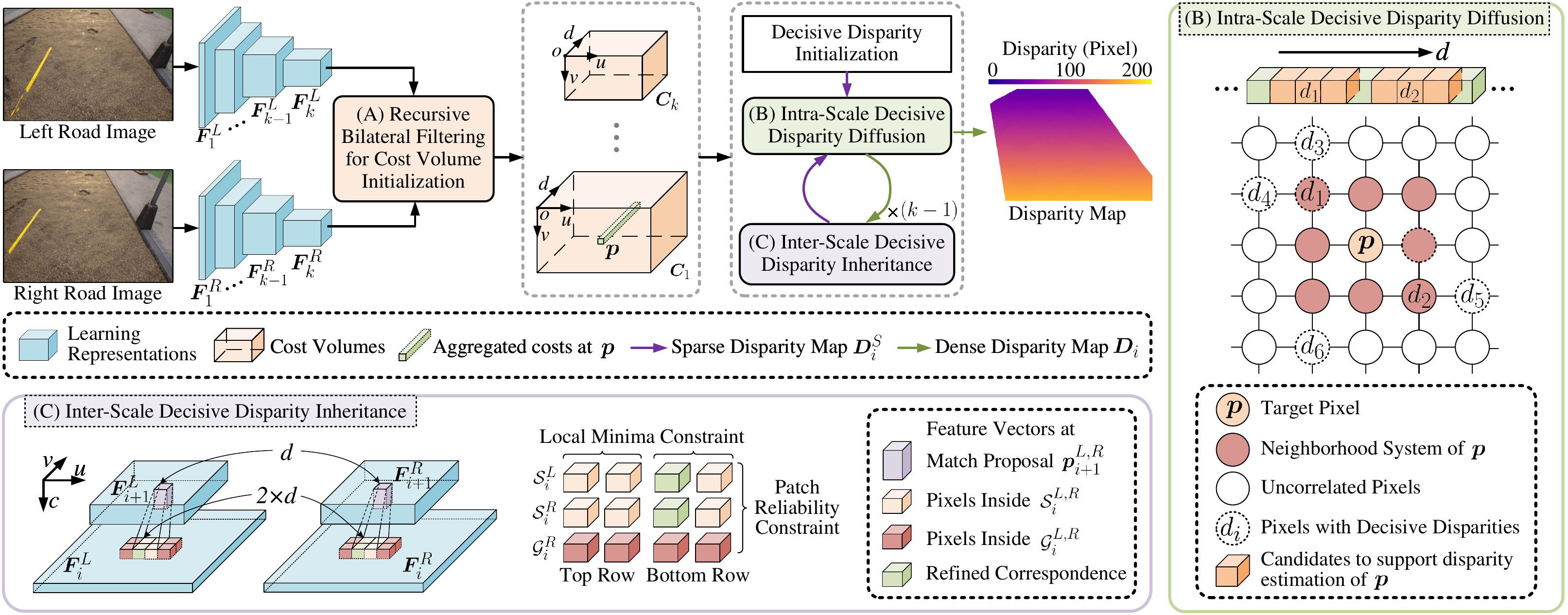}
		\centering
		\caption{An illustration of our proposed D3Stereo strategy. {Cost volume pyramid is first initialized with RBF. Afterwards, coarse decisive disparities initialized at the deepest layer are hierarchically propagated into former layers with alternating decisive disparity intra-scale diffusion and inter-scale inheritance algorithms.}}
		\label{fig.framework}
	\end{center}
\end{figure*}

\section{Methodology} 
\label{sec.methodology}
Based on Markov random field theory \cite{tappen2003comparison}, stereo matching can be formulated as an energy minimization problem \cite{fan2018road}:
\begin{equation}
E=\sum_{\boldsymbol{p}\in\boldsymbol{D}}^{}D(\boldsymbol{p},d)+\sum_{\boldsymbol{q}\in\mathcal{N}_{\boldsymbol{p}}}^{}V(\boldsymbol{p},\boldsymbol{q}),
\label{eq.E}
\end{equation}
where $\boldsymbol{p}=[u,v]^\top$ denotes a 2D pixel within the disparity map $\boldsymbol{D}$, function $D(\cdot)$ measures the stereo matching confidence at a given disparity $d$, function $V(\cdot)$ quantifies the compatibility between $\boldsymbol{p}$ and its neighborhood system $\mathcal{N}_{\boldsymbol{p}}$ comprising a collection of 2D pixels $\boldsymbol{q}$ adjacent to $\boldsymbol{p}$. As demonstrated in \cite{wang2021pvstereo}, confident disparities tend to have similar values, and their matching costs are consistent regardless of scales. Therefore, drawing inspiration from the scale-invariant feature detection introduced in \cite{lowe2004distinctive}, we extend $\mathcal{N}_{\boldsymbol{p}}$ to incorporate neighborhood systems of $\boldsymbol{p}$ across different scales, enabling pyramid stereo matching in this study. Following \cite{efe2021dfm}, we perform stereo matching via a hierarchical refinement strategy, as illustrated in Fig. \ref{fig.framework}. The process of $D(\cdot)$ is accomplished using either conventional explicit programming-based algorithms or pre-trained DCNN backbones, as detailed in Sect. \ref{sec.RBF}, while the process of $V(\cdot)$ is achieved through an intra-scale decisive disparity diffusion algorithm, and an inter-scale decisive disparity inheritance algorithm, as detailed in Sects. \ref{sec.intra} and \ref{sec.inter}, respectively. Our adopted sparse decisive disparity initialization approach obviates the necessity for additional keypoint detection and matching algorithms that are commonly used in conventional seed-and-grow stereo matching methods. Additionally, the combined use of decisive disparity intra-scale diffusion and inter-scale inheritance not only ensures the quantity and distribution uniformity of the estimated disparities but also significantly enhances stereo matching efficiency.

\subsection{Recursive Bilateral Filtering for Cost Volume Initialization}
\label{sec.RBF}
In our earlier research \cite{fan2018road}, we utilized the basic normalized cross-correlation (NCC) for stereo matching cost computation. Nevertheless, recent data-driven algorithms, generally developed based on DCNNs, have demonstrated superior performance, compared to such explicit programming-based methods. This is attributed to their capabilities of learning more {informative hierarchical representations} with pyramid structures, thereby offering a more effective solution for stereo matching challenges in complex scenarios. Hence in this paper, we develop D3Stereo for both explicit programming-based and {data-driven} methods. In this subsection, we detail only the cost volume initialization using pre-trained DCNN backbones (either the backbones pre-trained on the ImageNet \cite{russakovsky2015imagenet} database for natural image classification or those embedded in pre-trained deep stereo networks). Nevertheless, this procedure can also be accomplished through explicit block matching.

Given a pair of stereo road images $\boldsymbol{I}^{L}$ and $\boldsymbol{I}^{R}$, we first extract a collection of deep feature maps $\mathcal{F}^L=\{ \boldsymbol{F}_{1}^{L},\dots,\boldsymbol{F}_{k}^{L} \}$ and $\mathcal{F}^R=\{ \boldsymbol{F}_{1}^{R},\dots,\boldsymbol{F}_{k}^{R} \}$ at $k$ different resolutions using a pre-trained DCNN backbone. The feature maps $\boldsymbol{F}_{i}^{L,R}$ generally possess half the resolution of their shallower adjacent ones $\boldsymbol{F}_{i-1}^{L,R}$. A cost volume pyramid $\mathcal{C} = \{\boldsymbol{C}_1,\dots,\boldsymbol{C}_k\}$ can be subsequently obtained by computing the cosine similarity between each pair of left and right deep feature maps, respectively. The matching costs in $\mathcal{C}$ {undergo} normalization, with a lower matching cost indicating a better match. 

As a standard step in stereo matching algorithms, we conduct cost aggregation on the cost volumes to improve the piece-wise disparity coherency across the support region of each pixel \cite{zhang2017cross}. It has been mathematically proven that the function $V(\cdot)$ in (\ref{eq.E}) can be formulated through an adaptive cost aggregation process using a bilateral filter \cite{mozerov2015accurate}. A larger kernel size (commonly regarded as the ``receptive field'') often brings improved disparity estimation results. However, increasing the bilateral filtering kernel size can substantially lead to a notable increase in computational demands, thereby imposing significant memory pressure on parallel computing resources. 

A prevalent trend in network architecture design lies in replacing a large convolution kernel with stacked small ones \cite{simonyan2014very}. While possessing the same receptive field size, stacked small kernels exhibit lower computational complexity and greater network depth compared to a single large kernel. Motivated by this network architecture design, we introduce a recursive bilateral filtering algorithm for memory-efficient cost aggregation as follows:
\begin{equation}
\boldsymbol{C}^{(t)}_i(\boldsymbol{p},d)=\frac{\sum\limits_{\boldsymbol{q}\in   \mathcal{N}_{\boldsymbol{p},i}\cup\{ \boldsymbol{p} \} }{ \boldsymbol{K}_i(\boldsymbol{q})   \boldsymbol{C}}^{(t-1)}_i(\boldsymbol{p},d)}{ \sum\limits_{\boldsymbol{q}\in \mathcal{N}_{\boldsymbol{p},i}\cup\{ \boldsymbol{p} \} } \boldsymbol{K}_i(\boldsymbol{q})},
\label{eq.RBF}
\end{equation}
where $\boldsymbol{p}$ is a 2D pixel in $\boldsymbol{C}_{i}$, $d$ represents a disparity candidate at $\boldsymbol{p}$, ${\boldsymbol{C}}^{(t)}_i$ represents the $i$-th cost volume after the $t$-th RBF iteration with ${\boldsymbol{C}}^{(0)}_i={\boldsymbol{C}}_i$. In the RBF kernel:
\begin{equation}
\label{eq.kernel4}
\boldsymbol{K}_i(\boldsymbol{q})=\text{exp}\Big \{ - \frac{{   ||\boldsymbol{p}-\boldsymbol{q}||_2   }^2}{{\sigma_1}^{2}} - \frac{{(\boldsymbol{I}_i^{L}(\boldsymbol{p})-\boldsymbol{I}_i^{L}(\boldsymbol{q}))}^2}{{\sigma_2}^{2}} \Big \},
\end{equation}
$\mathcal{N}_{\boldsymbol{p},i}$ denotes a neighborhood system of $\boldsymbol{p}$ (the RBF kernel radius $\kappa_a=1$ corresponds to an eight-connected neighborhood system) at the $i$-th scale, $\boldsymbol{I}_i^{L}$ denotes a downsampled $\boldsymbol{I}^{L}$ with the same resolution as $\boldsymbol{F}_i^{L}$, $\sigma_1$ and $\sigma_2$ denote weighting parameters related to spatial distance and color similarity, respectively. As discussed in \cite{radosavovic2020designing}, executing $t_\text{max}$ iterations of bilateral filtering with a $3\times 3$ kernel is functionally equivalent in terms of receptive field size to performing the filtering process once, but with a $(2 t_\text{max} +1)\times (2 t_\text{max} +1)$ kernel. The computational consumption ratio of traditional bilateral filtering versus our proposed RBF is $\frac{1}{9}(4 t_\text{max}+\frac{1}{t_\text{max}}+4)$, which shows a monotonic increase {when $t_\text{max}>\frac{1}{2}$}. Moreover, it has been mathematically proven in \cite{luo2016understanding} that for a stack of convolutional layers, the weights of each pixel within its theoretical receptive field adhere to a Gaussian distribution. This concept naturally {translates to} the recursive structure of our proposed RBF. Therefore, with the same computational complexity, our proposed RBF can produce a larger receptive field adhering to a Gaussian distribution, thereby gathering more context information for cost aggregation. In addition, in practical implementations, the GPU memory needs are reduced by a factor of $\frac{1}{9}(4 t_\text{max}^2+4 t_\text{max}+1)$ when using our proposed RBF, significantly optimizing the memory resource usage. 

\subsection{Intra-Scale Decisive Disparity Diffusion}
\label{sec.intra}

As illustrated in Fig. \ref{fig.framework}, D3Stereo strategy is initialized with a collection of coarse decisive disparities, determined by measuring the peak ratio naive (PKRN) \cite{poggi2021confidence} scores and checking the left-right disparity consistency (LRDC) \cite{fan2018road} at the $k$-th layer, as employed in DFM \cite{efe2021dfm}. This process results in $\boldsymbol{D}_{k}^{S}$, a sparse disparity map with the lowest resolution. The linear hierarchical refinement structure employed in DFM has been proven to dramatically improve the stereo matching efficiency. However, the resulting disparity map is often quasi-dense, comprising a set of disparity clusters originating from a single initial decisive disparity. To address this issue, we introduce an intra-scale decisive disparity diffusion process positioned between two successive inter-scale refinement processes, denoted by an alternating hierarchical refinement structure (ARS). This novel contribution helps densify the sparse depth information initialized by the PKRN dense search process, thereby improving both the quantity and distribution uniformity of decisive disparities while retaining the efficiency gains achieved by the linear hierarchical refinement strategy in DFM. Our proposed intra-scale decisive disparity diffusion algorithm is developed based on the following hypotheses:
\begin{enumerate}[label=(\arabic*)]
    \item disparities change gradually across continuous regions;
    \item the matching cost of a desired disparity is a local minima;
    \item disparities between stereo images are consistent.
\end{enumerate}

We define a disparity state variable ${s}^{(t)}_{\boldsymbol{p},i}$ for $\boldsymbol{p}$ in the $t$-th iteration of decisive disparity diffusion when estimating the $i$-th disparity map $\boldsymbol{D}_i$ ($i\in [1,k] {\cap \mathbb{Z}}$) as follows: 
\begin{equation}
{s}^{(t)}_{\boldsymbol{p},i}=\mathop{\arg\min}_{s} \Big \{ 
\boldsymbol{C}_i(\boldsymbol{p},s) | s 
\in 
\Phi \big( \underset{\boldsymbol{q} \in \mathcal{N}_{\boldsymbol{p}, i}}{\bigcup} \big\{ \boldsymbol{D}_i^{(t-1)}(\boldsymbol{q})+ r \big\} \big)
\Big \},
\label{eq.d3_energy}
\end{equation}
where $r \in [-\tau, \tau] \cap \mathbb{Z}$ denotes the disparity search tolerance, in which $\tau \in \mathbb{Z}$ represents the disparity search bound, the neighborhood system $\mathcal{N}_{\boldsymbol{p},i}$ for disparity diffusion has a radius $\kappa_d$, $\Phi(\cdot)$ represents an operation to uniquify a given set, and $\boldsymbol{D}_i^{(t-1)}$ denotes the disparity map obtained after the $(t-1)$-th iteration with $\boldsymbol{D}_k^{(0)}=\boldsymbol{D}_k^S$, and $\boldsymbol{D}_i^{(0)}=\boldsymbol{D}_i^S$ ($i<k$) and are derived from the inter-scale refinement process at the $i$-th layer. The disparity state variable ${s}^{(t)}_{\boldsymbol{p},i}$ that fulfills the hypotheses (2) and (3) mentioned above is considered to be decisive. Afterwards, the newly generated decisive disparities are identically utilized to propagate disparity ranges to their neighborhood systems in the next iteration, corresponding to an ongoing process of depth information completion. To improve computational efficiency, we confine the intra-scale decisive disparity diffusion process only to the pixels whose neighborhood system has experienced state changes in the previous iteration. Moreover, unlike conventional unidirectional seed-growing process, we also incorporate an adversarial mechanism into our intra-scale decisive disparity diffusion process to update disparities that may have been determined incorrectly in the previous iterations. Specifically, if a pixel satisfies the following condition: 
\begin{equation}
\begin{split}
\mathop{\min} &\Big \{ 
\boldsymbol{C}_i(\boldsymbol{p},{s}^{(j)}_{\boldsymbol{p},i}) | j \in[0,t-2] \cap \mathbb{Z} 
\Big \}<
\\
 \mathop{\min} & \Big \{ 
\boldsymbol{C}_i(\boldsymbol{p},s) | s 
\in 
\Phi \big( \underset{\boldsymbol{q} \in \mathcal{N}_{\boldsymbol{p}, i}}{\bigcup} \big\{ \boldsymbol{D}_i^{(t-1)}(\boldsymbol{q})+ r \big\} \big)
\Big \},
\end{split}
\label{eq.adversarial}
\end{equation}
its disparity will be updated accordingly. This mechanism helps reduce the occurrence of incorrect disparities in the subsequent inter-scale decisive disparity inheritance process. Intra-scale decisive disparity diffusion terminates when no more pixels experience state changes (namely ${s}^{(t)}_{\boldsymbol{p},i}={s}^{(t-1)}_{\boldsymbol{p},i}$), resulting in a dense disparity map $\boldsymbol{D}_{i}$. Additional details on our proposed intra-scale decisive disparity diffusion strategy are provided in Algorithm \ref{alg.intra}. Subsequently, we follow our previous work \cite{fan2021rethinking} to perform perspective transformation using the $k$-th dense disparity map $\boldsymbol{D}_k$. By performing D3Stereo on the original left image and the transformed right image, the disparity search range is considerably narrowed and the similarity between matching blocks is further enhanced.

\begin{algorithm}[t!]
\small
\normalem
\setstretch{1.15}
\caption{Intra-Scale Decisive Disparity Diffusion}
\label{alg.intra}
\LinesNumbered 
\KwIn{Cost volume $\boldsymbol{C}_i$ and decisive disparity map $\boldsymbol{D}_i^S$}
\KwOut{Disparity map $\boldsymbol{D}_i$}
Initialize an empty set $\mathcal{P}$ to store the candidates for intra-scale decisive disparity diffusion\; 
$\boldsymbol{D}_i \gets \boldsymbol{D}_i^S$ \;
\For{$\boldsymbol{p}$ whose neighboring pixel $\boldsymbol{q}$ is determined to have a decisive disparity}{
        $\mathcal{P} \gets \mathcal{P} \cup \{\boldsymbol{p} \}$\;
}
\Repeat{no more pixels experience state changes}
{
    \For{$\boldsymbol{p} \in \mathcal{P}$}
    {
        $\mathcal{P} \gets \mathcal{P} - \{ \boldsymbol{p} \}$ \;
        Calculate its state ${s}^{(t)}_{\boldsymbol{p},i}$ using (\ref{eq.d3_energy})\;
        \If {its state satisfies hypotheses (2) and (3), or the adversarial mechanism condition in (\ref{eq.adversarial})} 
        {
            $\boldsymbol{D}_i(\boldsymbol{p}) \gets {s}^{(t)}_{\boldsymbol{p},i}$ and $\mathcal{P} \gets \mathcal{P} \cup \mathcal{N}_{\boldsymbol{p}}$\;
            }
    }
}
\end{algorithm}

\begin{figure}[t!]
    \centering
    \includegraphics[width=0.5\textwidth]{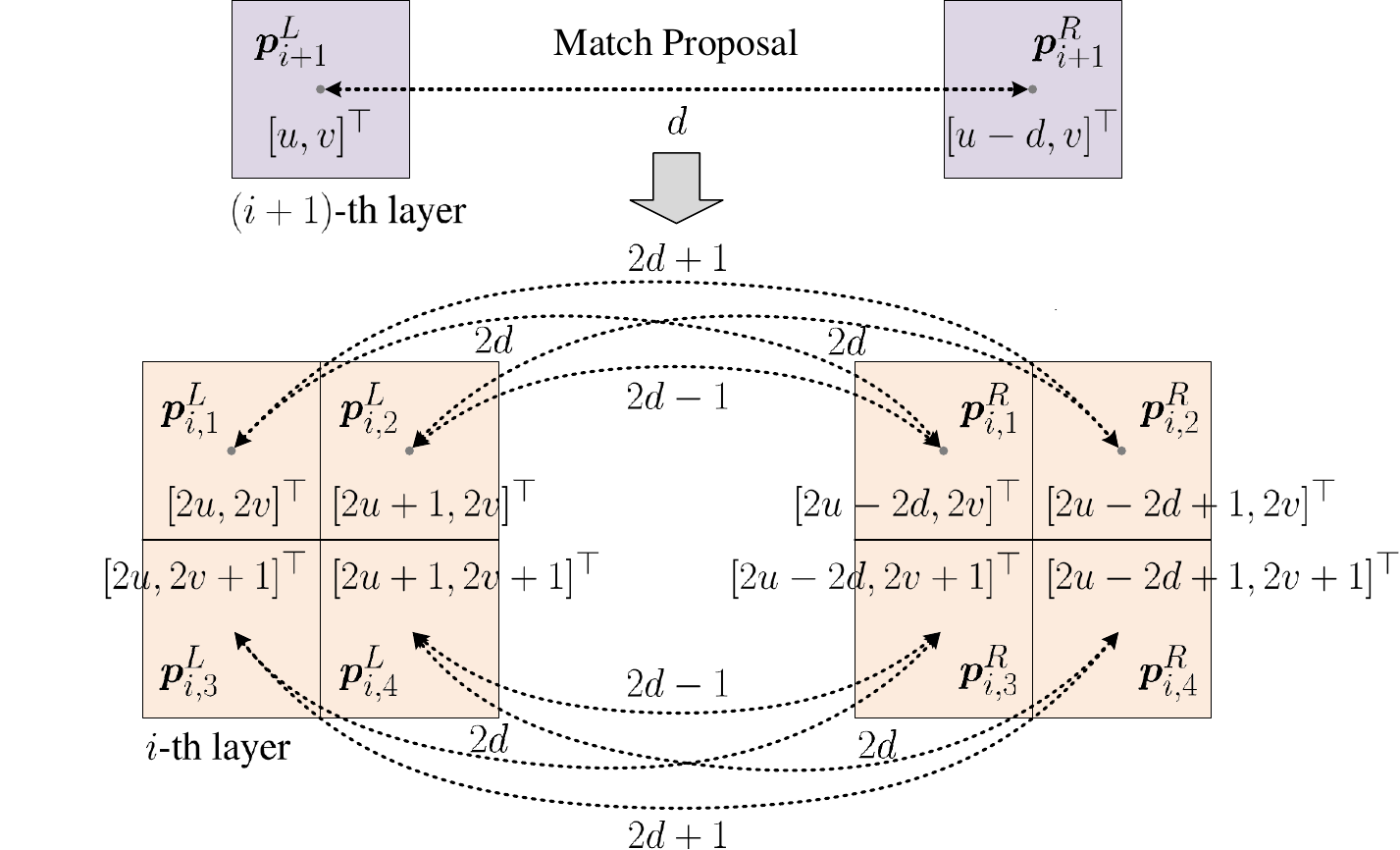}
    \caption{An illustration of the inter-scale decisive disparity inheritance process. {Each match proposal is mapped into a pair of patches with a size of $2\times 2$ pixels at the former layer, between which fine-grained matches are subsequently determined via PKRN and LRDC.}}
    \label{fig.patch}
\end{figure}

\subsection{Inter-Scale Decisive Disparity Inheritance}
\label{sec.inter}

The robustness of stereo matching using feature maps from deeper layers has been demonstrated in effectively addressing ambiguities, \egi, repetitive patterns and texture-less regions \cite{xu2023iterative}. This can be attributed to the richer semantic information within these feature maps generated through larger receptive fields \cite{zhou2023e3cm}. In contrast, feature maps from shallower layers focus on capturing local texture information and fine-grained details, and thus are more sensitive to pixel intensity changes. Shallower layers are advantageous for disparity estimation near/on discontinuities. They help in identifying small details, \egi, edges and surface textures \cite{fan2021rethinking}. Therefore, in this study, after estimating decisive disparities at the lowest resolution, we perform a series of inter-scale decisive disparity inheritance and intra-scale decisive disparity diffusion operations alternately, until we obtain a dense disparity map $\boldsymbol{D}_{1}$ at the highest resolution. This strategy incrementally introduces fine-grained details into the disparity map. 

In Patch2Pix \cite{zhou2021patch2pix}, each coarse match proposal $\mathcal{M}_{k}^{L,R}=\big\{\boldsymbol{p}_{k}^{L}, \boldsymbol{p}_{k}^{R} \big\}$ (determined by a decisive disparity in $\boldsymbol{D}^S_k$) are expanded into a pair of patches with a resolution of $2^{k-1}\times 2^{k-1}$ pixels at the $1$-st layer. All features from the layers $\{1,\dots,k\}$ that correspond to $\boldsymbol{p}_{1}^{L,R}$ within the patches are concatenated into a single feature vector, which is then fed to two regressors to determine the correspondence matching confidence. However, the simple concatenation operation may not fully exploit the fine-grained details present in the shallower layers, as features at deeper layers often have higher dimensions. To overcome this limitation, DFM \cite{efe2021dfm} employs a hierarchical refinement strategy to incrementally introduce fine-grained details into sparse correspondence matching. Specifically, in DFM, a pair of correspondences $\boldsymbol{p}_{i+1}^{L}$ and $\boldsymbol{p}_{i+1}^{R}$ matched at the $(i+1)$-th layer $(i < k)$ are mapped into a pair of patches $\mathcal{S}_{i}^{L,R}=\big\{\boldsymbol{p}_{i,1}^{L,R}, \dots, \boldsymbol{p}_{i,4}^{L,R} \big\}$ with a size of $2\times 2$ pixels at the $i$-th layer. Matches between $\mathcal{S}_{i}^{L,R}$ are subsequently determined via PKRN and LRDC, as introduced in Sect. \ref{sec.intra}. However, the PKRN-based hierarchical refinement constraint in DFM has two significant drawbacks: (1) determining satisfactory matches with PKRN and LRDC is not always effective and robust due to the limited matching candidates, which can lead to error accumulation and propagation from earlier stages to later stages; (2) PKRN requires a manually set threshold for each layer, which decreases its applicability and requires a manual algorithm tuning step.

This study focuses entirely on stereo matching, which is a 1D search problem. Therefore, the hierarchical refinement process aims at determining accurate matches between pixels in two co-row sets: $\mathcal{S}_{i,t}^{L,R}=\big\{ \boldsymbol{p}_{i,1}^{L,R}, \boldsymbol{p}_{i,2}^{L,R} \big\}$ and $\mathcal{S}_{i,b}^{L,R}=\big\{ \boldsymbol{p}_{i,3}^{L,R}, \boldsymbol{p}_{i,4}^{L,R} \big\}$, as illustrated in Fig. \ref{fig.patch}. The subscripts $t$ and $b$ denote the sets on the top and bottom rows, respectively. Following the intra-scale decisive disparity diffusion criteria stated in Sect. \ref{sec.intra}, we first analyze the reliability of the given patch pair based on the following hypotheses: 
\begin{enumerate}[label=(\arabic*)]
    \item disparities and matching costs within the patch are similar in continuous regions;
    \item the average of matching costs within the patch is lower than the minimum matching costs on the two sides of the patch, which is inherited from the match proposals that satisfy the local minima constraint. 
\end{enumerate}
We denote two sets that store the pixels on the left and right sides of $\mathcal{S}_{i,t}^{L,R}$ and $\mathcal{S}_{i,b}^{L,R}$ as $\mathcal{G}_{i,t}^{L,R}=\{ \boldsymbol{p}_{i,1}^{L,R}-[1,0]^\top, \boldsymbol{p}_{i,2}^{L,R}+[1,0]^\top \}$ and $\mathcal{G}_{i,b}^{L,R} =\{ \boldsymbol{p}_{i,3}^{L,R}-[1,0]^\top, \boldsymbol{p}_{i,4}^{L,R}+[1,0]^\top \}$, respectively, as visualized in Fig. \ref{fig.framework}. $\mathcal{S}_{i}^{L,R}$ is considered reliable when it satisfies the following patch reliability constraint (PRC): 
\begin{equation}
    \begin{split}
        &\mathrm{mean} \big\{ \Theta (\mathcal{S}_{i,t}^{L},\mathcal{S}_{i,t}^{R}), \Theta (\mathcal{S}_{i,b}^{L},\mathcal{S}_{i,b}^{R}) \big\}<
        \\
        &\min \big\{ \Theta(\mathcal{S}_{i,t}^{L},\mathcal{G}_{i,t}^{R}), \Theta(\mathcal{S}_{i,b}^{L},\mathcal{G}_{i,b}^{R}) \big\},
    \end{split}
    \label{eq.lecn}
\end{equation}
where 
\begin{equation}
\label{eq.lecm}
    \Theta (\mathcal{V}_i^L,\mathcal{V}_i^R) = \underset{\boldsymbol{p}_i^L \in\mathcal{V}^L_i}{\bigcup} 
\min \big\{ \boldsymbol{C}_{i}(\boldsymbol{p}_i^L,
    \| \boldsymbol{p}^L_i - \boldsymbol{p}^R_i \|_1
    ) | \boldsymbol{p}_i^R \in \mathcal{V}^R_i \big\},
\end{equation}
in which $\mathcal{V}_i^{L,R}$ denotes two input pixel sets. Otherwise, the given match proposal is considered unreliable and all pixels within $\mathcal{S}_{i}^{L,R}$ are abandoned. We then identify decisive disparities within the reserved $\mathcal{S}_{i}^{L,R}$, if it satisfies the following local minima constraint:
\begin{equation}
    \label{eq.extrema}
    \begin{split}
    \boldsymbol{C}_{i}(\boldsymbol{p}_i^L,d)<
    &\min 
    \big\{
    \big\{
    \boldsymbol{C}_{i}(\boldsymbol{p}_i^L,d+s)|s\in\{-1,1\}
    \big\} \cup 
    \\
    &\min \big\{ \Theta(\mathcal{S}_{i,t}^{L},\mathcal{G}_{i,t}^{R}), \Theta(\mathcal{S}_{i,b}^{L},\mathcal{G}_{i,b}^{R}) 
    \big\} 
    \big\}.
    \end{split}
\end{equation} 
The combined use of patch-based local minima constraint and search range propagation \cite{fan2018road} in (\ref{eq.extrema}) ensures a more critical determination for inter-scale decisive disparity inheritance, resulting in improved accuracy at the cost of reduced quantity. A sparse disparity map $\boldsymbol{D}_{i}^{S}$ is then obtained. It is noteworthy that both the patch reliability and local minima constraints are bidirectional as in the LRDC \cite{fan2018road}, and we only provide the left-to-right expressions in \eqref{eq.lecn} and \eqref{eq.extrema} for brevity. Additional details on our proposed inter-scale decisive disparity inheritance strategy are provided in Algorithm \ref{alg.cross}.

\begin{algorithm}[t!]
\small
\setstretch{1.25}
\caption{Inter-Scale Decisive Disparity Inheritance}
\label{alg.cross}
\LinesNumbered 
\KwIn{Cost volume $\boldsymbol{C}_{i}$ and disparity map $\boldsymbol{D}_{i+1}$}
\KwOut{Disparity map $\boldsymbol{D}_{i}^{S}$}
\For{$\boldsymbol{p}_{i+1}^{L,R}$ corresponding to a decisive disparity in $\boldsymbol{D}_{i+1}$}{
    Initialize $\mathcal{S}_{i}^{L,R}$ and $\mathcal{G}_{i}^{L,R}$\;
    \If {$\mathcal{S}_{i}^{L,R}$ and $\mathcal{G}_{i}^{L,R}$ satisfy the patch reliability constraint stated in \eqref{eq.lecn}} {
        \For {co-row pixels $\boldsymbol{p}_{i}^L\in\mathcal{S}_{i}^{L}$ and $\boldsymbol{p}_{i}^R\in\mathcal{S}_{i}^{R}$} {
            \If{$\boldsymbol{p}_{i}^L$ and $\boldsymbol{p}_{i}^R$ satisfy the local minima constraint stated in \eqref{eq.extrema}}{
                    $\boldsymbol{D}_{i}^S(\boldsymbol{p}^L_i) \gets \| \boldsymbol{p}^L_i - \boldsymbol{p}^R_i \|_1$
                }
        }
    }
}
\end{algorithm}

\section{Experimental Results}
\label{sec.experiments}
 
\subsection{Datasets, Implementation Details, and Evaluation Metrics}
Three datasets are used in our experiments: 
\begin{enumerate}[label=(\arabic*)]
    \item \textbf{UDTIRI-Stereo}: {Our proposed UDTIRI-Stereo dataset consists of 3,000 pairs of stereo images (resolution: 720 $\times$ 1,280 pixels), along with their disparity ground truth, collected across 12 scenarios under different illumination conditions (middle sunlight, intense sunlight, and street lighting at dark), weather conditions (tidy and watered), and road materials (asphalt and cement). To better simulate the roughness of actual roads, we introduce random 2D Perlin noise to the road data. In addition, we integrate digital twins of potholes acquired from the real world in our previously published work \cite{fan2018road} into the road data.}

    \item \textbf{Stereo-Road} \cite{fan2018road}: This dataset provides 71 pairs of well-rectified stereo road images (resolution: $609\times 1,240$ pixels), collected in Bristol, UK. As disparity ground truth is unavailable, we quantify the performance of stereo matching algorithms by measuring the similarity between the original left image and the warped right image generated using the estimated disparity maps.
    
    \item \textbf{Middlebury} \cite{scharstein2014high}: As our proposed D3Stereo strategy can also be employed to adapt any pre-trained DCNN to new scenarios, we further use this publicly available dataset (containing 15 pairs of stereo images along with their disparity ground truth) to evaluate the performance of the D3Stereo strategy in terms of general stereo matching, where substantial discontinuities are present.
    
\end{enumerate}

We first validate the effectiveness of adapting stereo matching DCNNs (without any model fine-tuning), including PSMNet \cite{PSMNet}, AANet \cite{AA}, BGNet \cite{BGNet}, LacGwc \cite{LacGwc}, GMStereo \cite{Unimatch}, CreStereo \cite{Cre}, and two domain generalization-aimed networks, GraftNet \cite{liu2022graftnet} and HVTStereo \cite{chang2023domain}, pre-trained on the KITTI Stereo dataset \cite{menze2015object} and SceneFlow dataset \cite{mayer2016large}, using our proposed D3Stereo strategy for road disparity estimation. Additionally, we compare our proposed D3Stereo matching algorithm (abbreviated as PT-D3Stereo), which employs NCC along with PT for cost volume pyramid construction, with three state-of-the-art (SoTA) explicit programming-based stereo matching algorithms: PT-SRP \cite{fan2018road}, PT-FBS \cite{FBS}, and PT-SGM \cite{fan2021rethinking}, developed specifically for road surface 3D reconstruction. The above two experiments were conducted on the UDTIRI-Stereo and Stereo-Road datasets. Moreover, we employ the same experimental setups to conduct additional experiments on the Middlebury dataset to further demonstrate the effectiveness of our proposed D3Stereo strategy for general stereo matching. Finally, D3Stereo is applied to backbone DCNNs pre-trained on the ImageNet database for demonstrating its compatibility with general-propose backbone DCNNs. All experiments were conducted on an NVIDIA RTX 4090 GPU and an Intel i7-12700K CPU.

Since the UDTIRI-Stereo and Middlebury datasets provide disparity ground truth, we adopt end-point error (EPE) and percentage of error pixels (PEP) with a tolerance of $\delta$ pixels for performance quantification. For experiments conducted on the Stereo-Road dataset, we first warp the right images into the left view using the estimated disparity maps, and then calculate the structure similarity index measure (SSIM), mean squared error (MSE), and peak signal-to-noise ratio (PSNR) metrics between the original left images and generated images to quantify the accuracy of the stereo matching algorithms. 

\subsection{Hyperparameter Selection in Decisive Disparity Diffusion}
\label{sec.ep.abla_d3}

\begin{figure}[t!]
    \begin{center}
        \centering
        \includegraphics[width=0.5\textwidth]{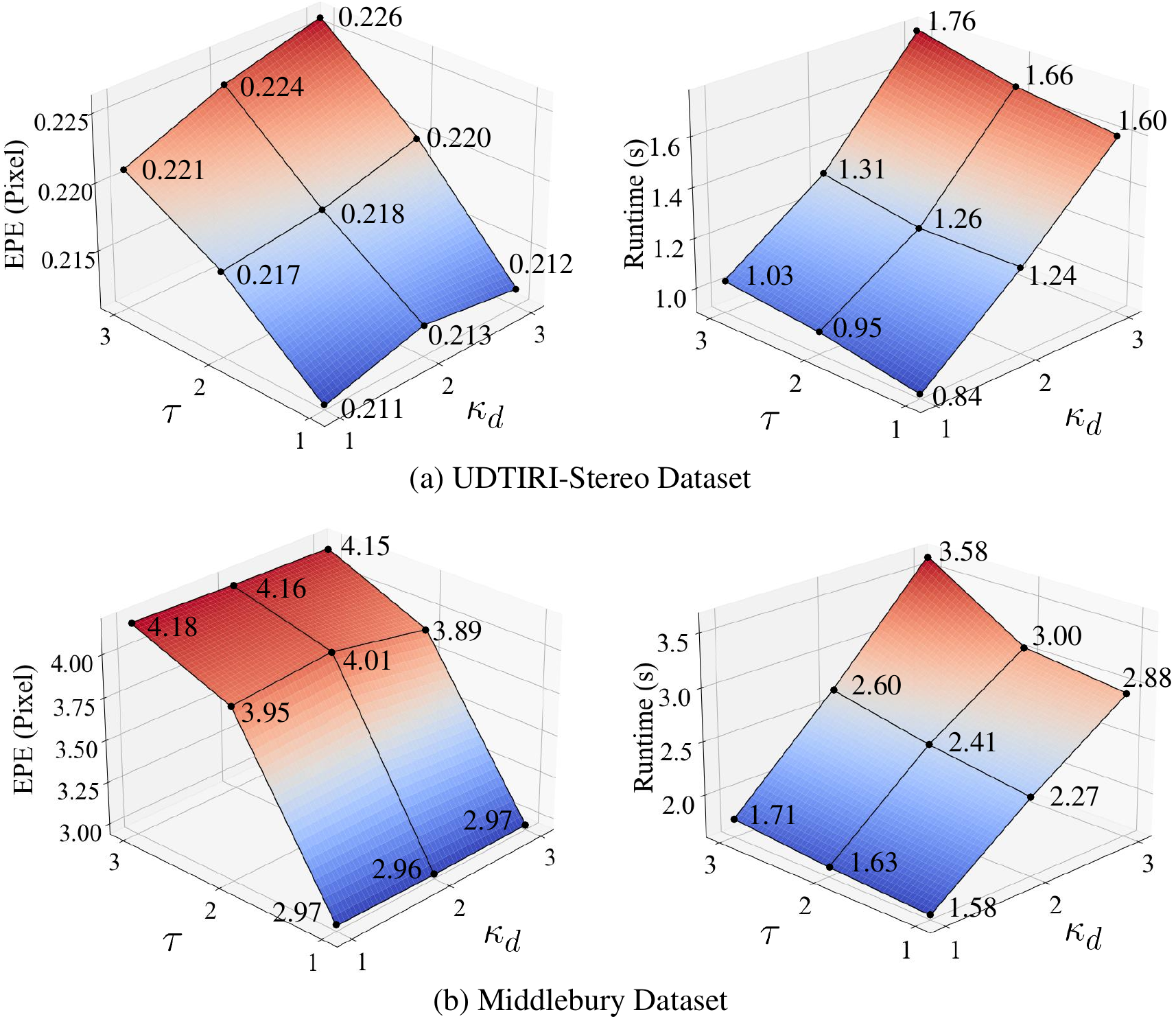}
        \centering
        \caption{Experimental results regarding hyperparameter selection for decisive disparity diffusion.}
        \label{fig.SR}
    \end{center}
\end{figure}

We provide details on the selection of disparity search bound $\tau$ and diffusion neighborhood radius $\kappa_{d}$ used in the intra-scale decisive disparity diffusion process. The EPE and runtime of stereo matching with respect to different $\tau$ and $\kappa_{d}$ are given in Fig. \ref{fig.SR}. It can be observed that setting $\tau=\kappa_{d}=1$ results in the best overall performance in stereo matching, and increasing $\tau$ and $\kappa_{d}$ leads to a noticeable increase in both the EPE and runtime. This phenomenon can be attributed to the introduction of more unreliable disparity candidates, when using higher $\tau$ and $\kappa_{d}$ during decisive disparity diffusion. Moreover, the weighting parameters $\sigma_{1,2}$ and the maximum iteration $t_\text{max}$ in our RBF are determined through a brute-force search on the Middlebury 2014 dataset \cite{scharstein2014high}. 

\begin{figure}[t!]
\centering
\includegraphics[width=0.5\textwidth]{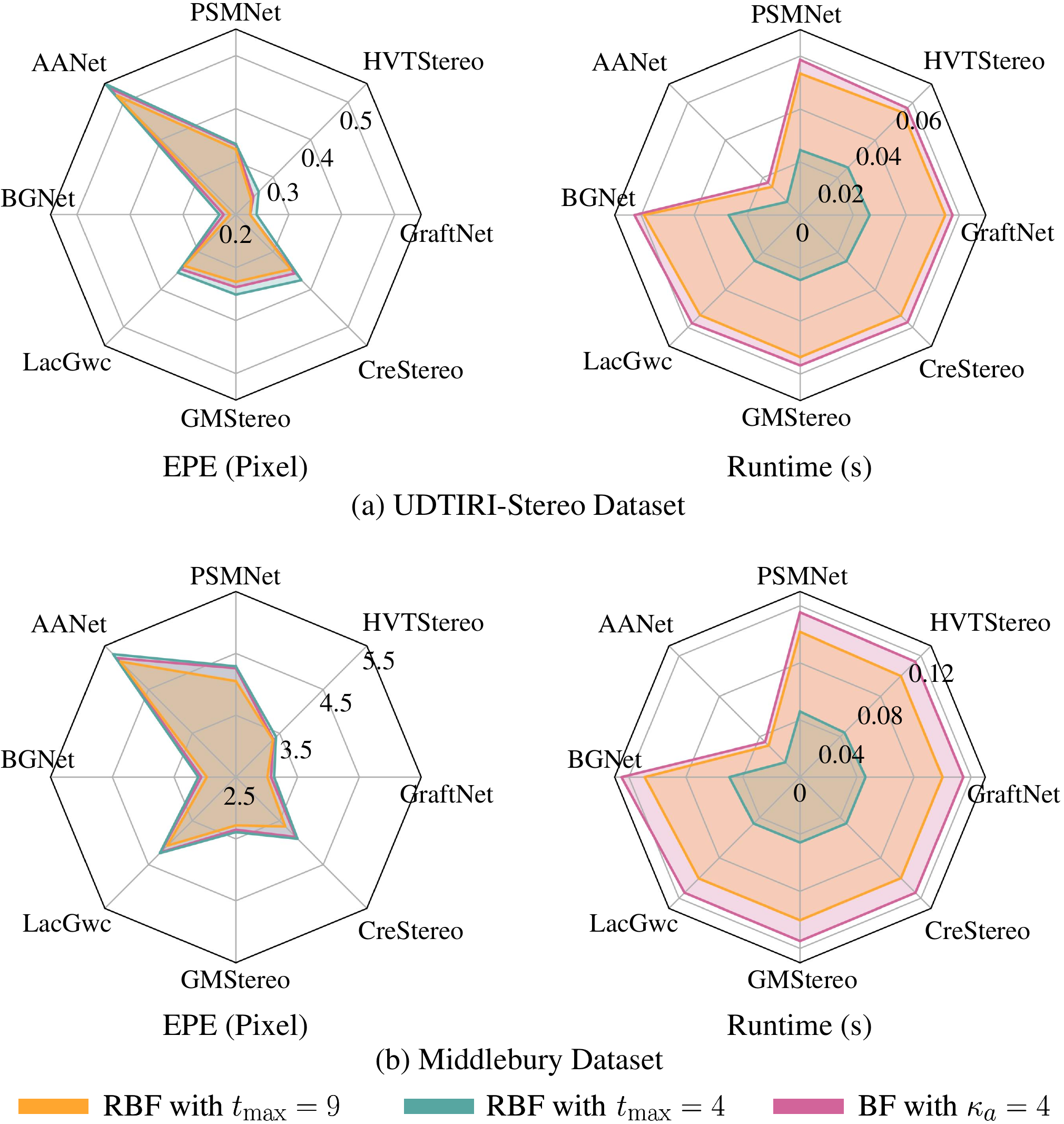}
\centering
\caption{{Comparisons between RBF and RF when having identical computational complexity or an identical receptive field.}}
\label{fig.RBF}
\end{figure}

\begin{table}[t!]
    \fontsize{7}{10}\selectfont
    \caption{Ablation study on the essential components of D3Stereo strategy. $\downarrow$ represents that lower values correspond to better performance. $\uparrow$ represents that higher values correspond to better performance. The best results are shown in bold type.}
    \setlength{\tabcolsep}{1mm}
    \label{tab.Abla}
    \begin{subtable}[t]{0.5\textwidth}
        \centering
        \caption{Experimental results on the UDTIRI-Stereo dataset.}
        \setlength{\tabcolsep}{0.8mm}{
    	\begin{tabular}
    		{cccccccc}
    		\toprule[1.5pt]
    		\multirow{2}{*}{PT} & \multirow{2}{*}{PRC} & \multirow{2}{*}{RBF} & \multirow{2}{*}{ARS} & \multicolumn{2}{c}{PEP (\%) $\downarrow$} & \multirow{2}{*}{EPE (pixel)$\downarrow$} & \multirow{2}{*}{Runtime (s)$\downarrow$}\\
    		\specialrule{0em}{1pt}{1pt}
    		\cline{5-6}
    		\specialrule{0em}{1pt}{1pt}
    		&&&&$\delta$=0.5&$\delta$=1&\\
    		\midrule[1pt]
            & & & & 10.5 & 3.57  & 0.48 & {0.79} \\  
            \ding{52}& & & & 9.15 & 3.27  & 0.38 & 0.62 \\ 
            \ding{52}& \ding{52}& & & 9.13 & 3.22 & 0.37 & 0.64 \\ 
            \ding{52}& &\ding{52}& & 9.22 & 2.51 & 0.29 & 0.91 \\ 
            \ding{52}& & &\ding{52}& 9.60 & 3.64 & 0.44 & \textbf{0.59} \\ 
            \ding{52}& \ding{52}&\ding{52}& & 4.54 & 2.23 & 0.28 & 0.94 \\
            \ding{52}& &\ding{52}&\ding{52}& 3.64 & 1.37 & 0.26 & 0.90 \\
            \ding{52}& \ding{52}& &\ding{52}& 8.31 & 2.75 & 0.35 & 0.61 \\     
            \ding{52}& \ding{52}&\ding{52}&\ding{52}& \textbf{3.53} & \textbf{1.20} & \textbf{0.21} & 0.84 \\ 
        	\bottomrule [1.5pt]   	
    	\end{tabular}}
    \vspace{12pt}
    \label{tab.Abla_a}
    \end{subtable}
    \begin{subtable}[t]{0.5\textwidth}
        \centering
        \caption{Experimental results on the Middlebury dataset.}
        \setlength{\tabcolsep}{0.9mm}{
        \begin{tabular}{ccccccc}
            \toprule[1.5pt]
            \multirow{2}{*}{PRC\hspace{4pt}} & \multirow{2}{*}{RBF} & \multirow{2}{*}{ARS\hspace{4pt}} & \multicolumn{2}{c}{PEP (\%) $\downarrow$} & \multirow{2}{*}{EPE (pixel)$\downarrow$} & \multirow{2}{*}{Runtime (s) $\downarrow$}\\
            \specialrule{0em}{1pt}{1pt}
            \cline{4-5}
            \specialrule{0em}{1pt}{1pt}
            &&&$\delta$=1&$\delta$=2&\\
            \midrule[1pt]
             & & & 34.3 & 24.0 & 4.07 & 1.11 \\  
            \ding{52}& & & 34.2 & 23.8 & 3.75 & 1.03 \\ 
            &\ding{52}& & 25.2 & 17.6 & 3.41 & 1.89 \\ 
            & &\ding{52}& 34.4 & 23.8 & 4.17 & \textbf{0.94} \\ 
            \ding{52}&\ding{52}& & 25.2 & 17.4 & 3.15 &1.90 \\ 
            &\ding{52}&\ding{52}& 25.4 & 17.8 & 3.46 & 1.77 \\     
            \ding{52}& &\ding{52}& 33.1 & 22.8 & 3.39 & 0.96 \\  
            \ding{52}&\ding{52}&\ding{52}& \textbf{24.9}  & \textbf{17.1} & \textbf{2.97} &1.58 \\  
            \bottomrule [1.5pt]   	 
	    \end{tabular}}
        \label{tab.Abla_b}
    \end{subtable}
\end{table}

\subsection{Ablation Study}

We first conduct an ablation study to evaluate the cost aggregation efficiency between RBF and BF, and their impacts on stereo matching accuracy. As discussed in Sect. \ref{sec.RBF}, BF with $\kappa_{a}=4$ has an identical receptive field compared to RBF with $t_\text{max}=4$, and has identical theoretical computational complexity compared to RBF with $t_\text{max}=9$. Therefore, we compare the performance of eight stereo matching networks using RBF and BF with these three parameter settings, as presented in Fig. \ref{fig.RBF}. It can be observed that with an identical receptive field, our RBF with $t_\text{max}=4$ witnesses significantly improved cost aggregation efficiency while leading to slightly decreased stereo matching accuracy, compared to BF with $\kappa_{a}=4$. We attribute this accuracy gap to the low effectiveness of small filtering kernels in RBF to aggregate matching costs between long-distance pixels. Specifically, the small kernels are unable to establish direct interactions between pixels with similar disparities and intensities but at long distances, thus making RBF less effective in areas with intense texture variations compared to BF. However, with identical theoretical computational complexity, BRF with $t_\text{max}=9$ exhibits superiority in both cost aggregation efficiency and stereo matching accuracy. These improvements can be attributed to RBF's recursive filtering process, which expands the receptive field to aggregate matching costs from more related pixels. Additionally, the smaller filtering kernel in RBF results in decreased computations in the weighting initialization process compared to BF. In general, our proposed RBF strikes a better balance between stereo matching accuracy and cost aggregation efficiency compared to BF.

We conduct another ablation study to determine the essential components of D3Stereo strategy. {The baseline setup includes the PKRN-based hierarchical refinement constraint and linear hierarchical refinement structure used in the DFM \cite{efe2021dfm}, and the traditional BF}. As shown in Table \ref{tab.Abla}, D3Stereo with all these components achieves the highest stereo matching accuracy on both datasets. Secondly, as expected, PT significantly improves both the stereo matching accuracy and efficiency, which is consistent with the findings in \cite{fan2018road}. Additionally, RBF improves stereo matching when combined with any of the components. Specifically, EPE decreases by 23.7-40.9\% on our UDTIRI-Stereo dataset and 5.7-16.2\% on the Middlebury dataset. Although RBF increases the runtime of D3Stereo by over 50\% on the Middlebury dataset, this increase is lower to 35\% on our UDTIRI-Stereo dataset when used in conjunction with PT, which significantly reduces the stereo matching search range. Finally, the collaboration between our introduced PRC and the ARS leads to improvements in both the stereo matching accuracy and efficiency. However, when used alone, they either improve only disparity estimation accuracy or efficiency while decreasing the other. 

\begin{table*}[t!]
    \fontsize{7}{10}\selectfont 
    \caption{Experimental results on the UDTIRI-Stereo dataset. The best results are shown in bold type.}
    \setlength{\tabcolsep}{1mm}
    \label{tab.SYN1}
    \begin{subtable}[t]{1\textwidth}
        \centering
        \caption{Comparison among SoTA explicit programming-based disparity estimation algorithms developed specifically for road surface 3D reconstruction.}
        \begin{tabular}{lcccccc}
        \toprule[1.5pt]
        \multirow{2}{*}{Method\hspace{50pt}} & \multicolumn{2}{c}{PEP (\%) $\downarrow$} & \multirow{2}{*}{\makebox[0.09\textwidth][c]{EPE (pixel) $\downarrow$}} & \multirow{2}{*}{\makebox[0.1\textwidth][c]{PSNR (dB) $\uparrow$}} & \multirow{2}{*}{\makebox[0.1\textwidth][c]{MSE $\downarrow$}} & \multirow{2}{*}{\makebox[0.1\textwidth][c]{SSIM $\uparrow$}}\\
        \specialrule{0em}{1pt}{0pt}
        \cline{2-3}
        \specialrule{0em}{1pt}{1pt}
        &\makebox[0.09\textwidth][c]{$\delta$=0.5}&\makebox[0.09\textwidth][c]{$\delta$=1}&&\\
        \midrule[1pt]
        PT-SRP \cite{fan2018road}& 43.2 & 33.0 & 1.27 & 33.01 & 52.12 & 0.867 \\            
        PT-FBS \cite{FBS}& 33.7 & 10.2 & 0.75 & 32.87 & 48.80 & 0.895 \\
        PT-SGM \cite{fan2021rethinking}& 7.17 & 4.13 & 0.72 & 32.50 & \textbf{41.12} & 0.932 \\  
        \textbf{PT-D3Stereo (Ours)}& \textbf{5.90} & \textbf{2.99} & \textbf{0.65} & \textbf{33.14} & 42.58 & \textbf{0.951} \\   
        \bottomrule [1.5pt]   	
    \vspace{12pt}
    \end{tabular}
    \label{tab.SYN1_a}
    \end{subtable}
    \begin{subtable}[t]{1\textwidth}
        \centering
        \caption{Comparisons of SoTA stereo matching networks without and with our proposed D3Stereo strategy applied.}
        \begin{tabular}{lcccccc}
            \toprule[1.5pt]
            \multirow{2}{*}{Method} & \multicolumn{2}{c}{PEP (\%) $\downarrow$} & \multirow{2}{*}{\makebox[0.09\textwidth][c]{EPE (pixel) $\downarrow$}} & \multirow{2}{*}{\makebox[0.09\textwidth][c]{PSNR (dB) $\uparrow$}} & \multirow{2}{*}{\makebox[0.09\textwidth][c]{MSE $\downarrow$}} & \multirow{2}{*}{\makebox[0.09\textwidth][c]{SSIM $\uparrow$}}\\
            \specialrule{0em}{1pt}{0pt}
            \cline{2-3}
            \specialrule{0em}{1pt}{1pt}
            &\makebox[0.09\textwidth][c]{$\delta$=0.5}&\makebox[0.09\textwidth][c]{$\delta$=1}&&\\
            \midrule[1pt]
            PSMNet \cite{PSMNet}& 46.7 & 13.2 & 0.84 & 32.36 & 55.51 & 0.894 \\
            \textbf{PSMNet+D3Stereo (Ours)} & \textbf{4.81} & \textbf{2.21} & \textbf{0.31} & \textbf{34.75} & \textbf{32.80} & \textbf{0.953} \\
            \specialrule{0em}{1pt}{0pt} 
            \cline{1-7}
            \specialrule{0em}{1pt}{1pt}
            AANet \cite{AA} & 35.4 & 8.69 & 0.56 & 34.01 & 38.14 & 0.932 \\
            \textbf{AANet+D3Stereo (Ours)} & \textbf{11.1} & \textbf{1.79} & \textbf{0.43} & \textbf{34.21} & \textbf{36.49} & \textbf{0.948} \\
            \specialrule{0em}{1pt}{0pt} 
            \cline{1-7}
            \specialrule{0em}{1pt}{1pt}
            BGNet \cite{BGNet}& 12.7 & 1.51 & 0.26 & 34.59 & 36.10 & 0.948 \\
            \textbf{BGNet+D3Stereo (Ours)}& \textbf{3.53} & \textbf{1.20} & \textbf{0.21} & \textbf{34.72} & \textbf{34.07} & \textbf{0.954} \\            
            \specialrule{0em}{1pt}{0pt} 
            \cline{1-7}
            \specialrule{0em}{1pt}{1pt}
            LacGwc \cite{LacGwc}& 18.6 & 2.87 & \textbf{0.36} & 34.45 & 36.91 & 0.945 \\
            \textbf{LacGwc+D3Stereo (Ours)} & \textbf{4.61} & \textbf{2.01} & 0.37 & \textbf{34.61} & \textbf{34.28} & \textbf{0.953} \\
            \specialrule{0em}{1pt}{0pt} 
            \cline{1-7}
            \specialrule{0em}{1pt}{1pt}
            GMStereo \cite{Unimatch}& 23.6 & 4.12 & 0.43 & 32.66 & 47.39 & 0.937 \\
            \textbf{GMStereo+D3Stereo (Ours)}& \textbf{3.91} & \textbf{1.46} & \textbf{0.31} & \textbf{34.66} & \textbf{33.48} & \textbf{0.953} \\
            \specialrule{0em}{1pt}{0pt} 
            \cline{1-7}
            \specialrule{0em}{1pt}{1pt}
            CreStereo \cite{Cre}& 6.02 & \textbf{1.41} & 0.40 & \textbf{34.77} & 34.85 & 0.950 \\
            \textbf{CreStereo+D3Stereo (Ours)}& \textbf{4.58} & {1.50} & \textbf{0.34} & 34.60 & \textbf{33.87} & \textbf{0.952}  \\
            \specialrule{0em}{1pt}{0pt} 
            \cline{1-7}
            \specialrule{0em}{1pt}{1pt}
            GraftNet \cite{liu2022graftnet}& 20.9 & 2.78 & 0.33 & 34.72 & 36.69 & 0.943 \\
            \textbf{GraftNet+D3Stereo (Ours)}& \textbf{4.25} & \textbf{1.34} & \textbf{0.22} & \textbf{35.17} & \textbf{32.35} & \textbf{0.951}   \\
            \specialrule{0em}{1pt}{0pt} 
            \cline{1-7}
            \specialrule{0em}{1pt}{1pt}
            HVTStereo \cite{chang2023domain}& 6.83 & 1.95 & 0.36 & 34.75 & 35.65 & 0.937 \\
            \textbf{HVTStereo+D3Stereo (Ours)}& \textbf{4.91} & \textbf{1.67} & \textbf{0.26} & \textbf{34.79} & \textbf{31.11} & \textbf{0.951}  \\
            \bottomrule [1.5pt]   	
        \end{tabular}
        \label{tab.SYN1_b}
    \end{subtable}
\end{table*}

\begin{figure}[t!]
	\centering
	\includegraphics[width=0.5\textwidth]{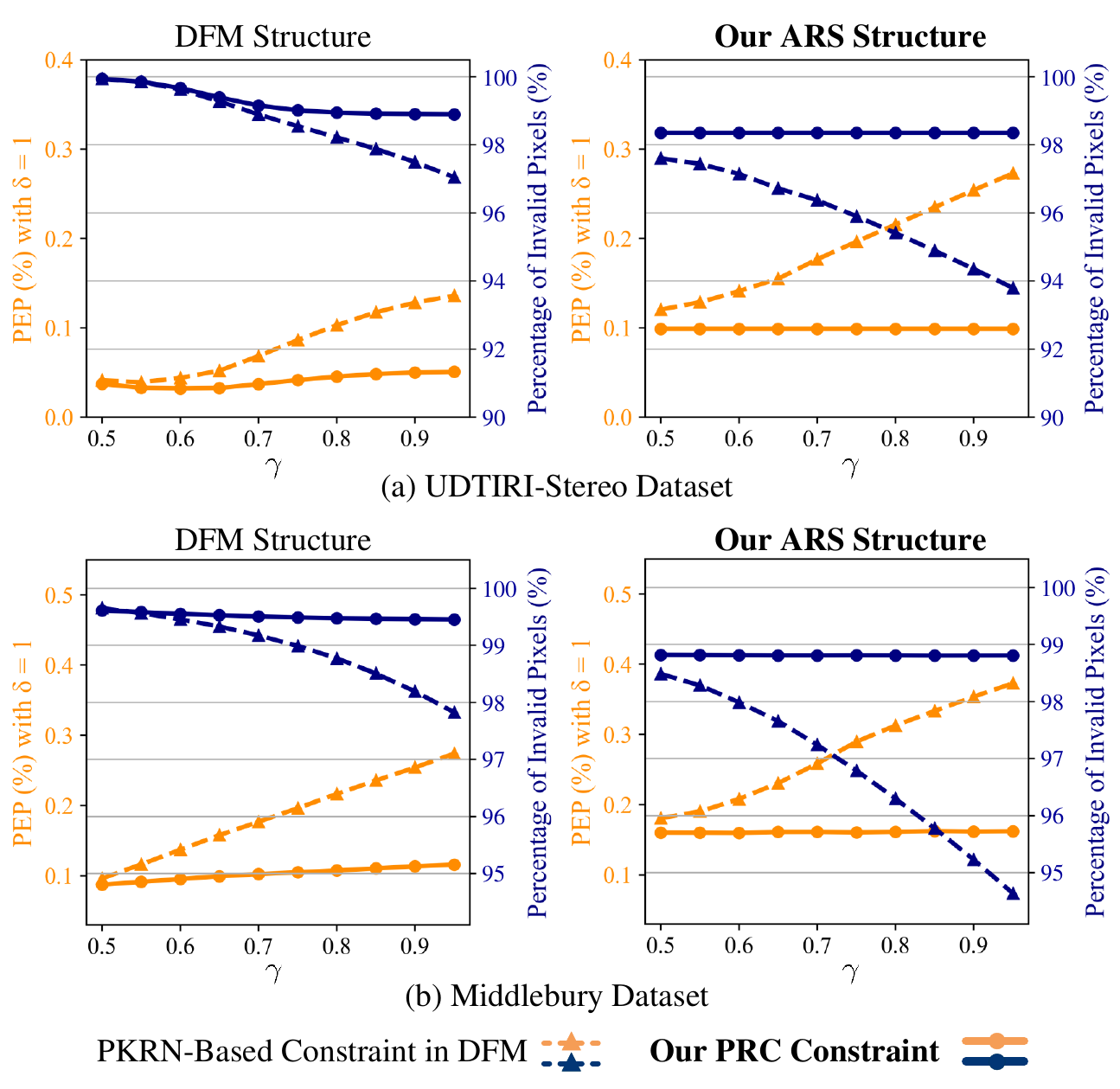}
	\centering
        \caption{Quantitative results of $\boldsymbol{D}^S_1$ yielded with different hierarchical refinement structures and constraints.}
	\label{fig.abla}
\end{figure}

\begin{figure*}[t!]
\centering
    \includegraphics[width=1\textwidth]{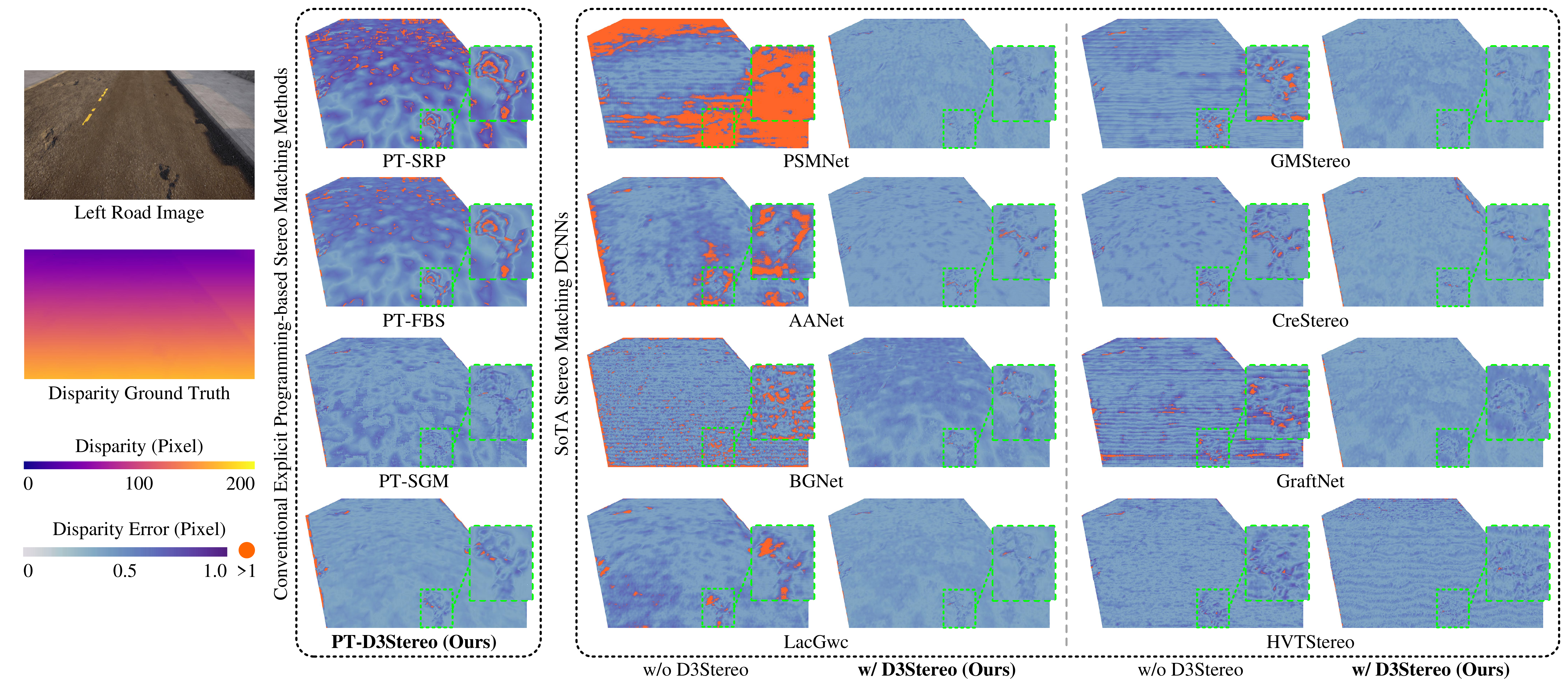}
    \caption{Examples of disparity estimation results on our created UDTIRI-Stereo dataset.}
    \label{fig.vis_syn1}
\end{figure*}

We further validate the effectiveness of our introduced PRC and the alternating hierarchical refinement structure. The percentage of error pixels and invalid pixels of decisive disparities in $\boldsymbol{D}^S_1$ yielded using different hierarchical refinement strategies are provided in Fig. \ref{fig.abla}, where $\gamma$ represents the PKRN threshold in the decisive disparity initialization process at the $k$-th layer. Our observations indicate that the alternating structure considerably increases the quantity of decisive disparities while leading to decreased accuracy. In contrast, our introduced PRC significantly enhances the accuracy of the decisive disparities, albeit with a decrease in their quantity. However, when these two components are used jointly, both the quantity and reliability of decisive disparities are significantly improved. While the quantity of decisive disparities obtained using our alternating hierarchical refinement strategy is lower than that achieved by DFM, our structure ensures a more uniform distribution of decisive disparities, resulting in improvements in both disparity estimation accuracy and efficiency in D3Stereo. Moreover, our hierarchical refinement strategy is significantly less sensitive to the impact of $\gamma$, demonstrating better adaptivity, compared to DFM.

\subsection{Comparisons with SoTA Algorithms for Road Surface 3D Reconstruction}
\label{sec.ep.stereo}

\begin{figure*}[t!]
\centering
    \includegraphics[width=1\textwidth]{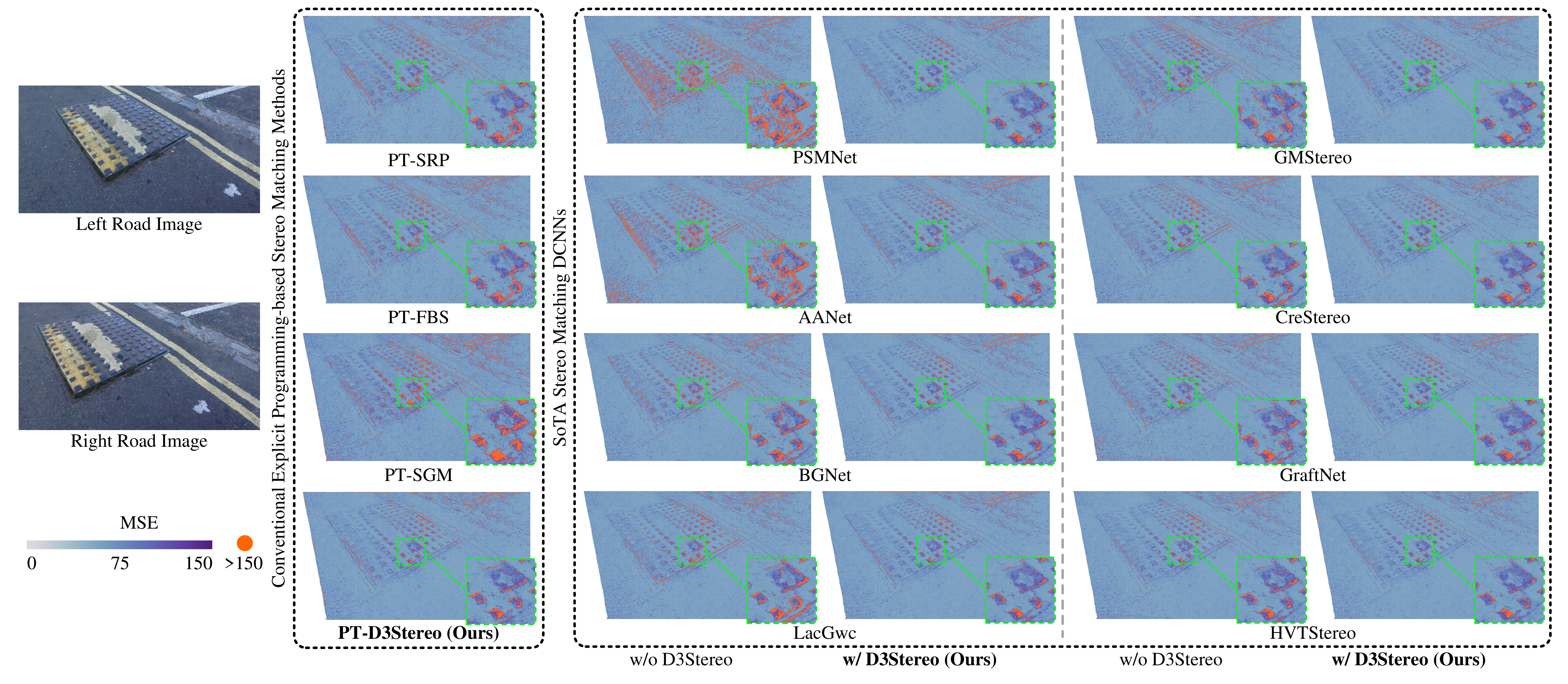}
    \caption{Examples of stereo image reconstruction results on the Stereo-Road dataset.}
    \label{fig.vis_real1}
\end{figure*}

\begin{table}[t!]
    \fontsize{7}{10}\selectfont 
    \caption{Experimental results on the Stereo-Road dataset \cite{fan2018road}. The best results are shown in bold type.}
    \setlength{\tabcolsep}{1.5mm}
    \label{tab.Real1}
    \begin{subtable}[t]{0.5\textwidth}
    \centering
    \caption{Comparison among SoTA explicit programming-based disparity estimation algorithms developed specifically for road surface 3D reconstruction.}
        \begin{tabular}
            {lccc}
    	\toprule[1.5pt]
    	{Method\hspace{50pt}} & \makebox[0.13\textwidth][c]{PSNR (dB) $\uparrow$} & \makebox[0.13\textwidth][c]{MSE $\downarrow$} & \makebox[0.13\textwidth][c]{SSIM $\uparrow$} \\
    	\midrule[1pt]
            PT-SRP \cite{fan2018road}& 30.82 & 68.91 & 0.909 \\
            PT-FBS \cite{FBS}& 30.83 & 68.14 & 0.911 \\
            PT-SGM \cite{fan2021rethinking}& 30.08 & 74.80 & 0.906 \\
            \textbf{PT-D3Stereo (Ours)}& \textbf{31.00} & \textbf{63.37} & \textbf{0.930} \\            
            \bottomrule [1.5pt]   
    \vspace{12pt}
        \end{tabular}
    \label{tab.Real1_a}
    \end{subtable}
    \begin{subtable}[t]{0.5\textwidth}
        \centering
        \caption{Comparisons of SoTA stereo matching networks without and with our proposed D3Stereo strategy applied.}
        \begin{tabular}{lccc}
            \toprule[1.5pt]
            {Method} & \makebox[0.13\textwidth][c]{PSNR (dB) $\uparrow$} & \makebox[0.13\textwidth][c]{MSE $\downarrow$} & \makebox[0.13\textwidth][c]{SSIM $\uparrow$} \\
            \midrule[1pt]
            PSMNet \cite{PSMNet}& 28.68 & 122.5 & 0.829 \\
            \textbf{PSMNet+D3Stereo (Ours)}& \textbf{31.73} & \textbf{55.15} & \textbf{0.938} \\
            \hline
		  \specialrule{0em}{1pt}{0pt}
            AANet \cite{AA}& 29.79 & 90.49 & 0.885 \\
            \textbf{AANet+D3Stereo (Ours)}& \textbf{31.34} & \textbf{60.50} & \textbf{0.931} \\
            \hline
		  \specialrule{0em}{1pt}{0pt}
            BGNet \cite{BGNet}& 31.46 & 59.67 & 0.928 \\
            \textbf{BGNet+D3Stereo (Ours)}& \textbf{31.65} & \textbf{56.17} & \textbf{0.937} \\
            \hline
		  \specialrule{0em}{1pt}{0pt}
            LacGwc \cite{LacGwc}& 31.32 & 62.02 & 0.923 \\
            \textbf{LacGwc+D3Stereo (Ours)}& \textbf{31.70} & \textbf{55.43} & \textbf{0.938} \\
            \hline
		  \specialrule{0em}{1pt}{0pt}
            GMStereo \cite{Unimatch}& 30.98 & 67.10 & 0.915 \\
            \textbf{GMStereo+D3Stereo (Ours)}& \textbf{31.61} & \textbf{56.34} & \textbf{0.937} \\
            \hline
		  \specialrule{0em}{1pt}{0pt}
            CreStereo \cite{Cre}& 31.44 & 60.25 & 0.927 \\
            \textbf{CreStereo+D3Stereo (Ours)}& \textbf{31.57} & \textbf{57.05} & \textbf{0.936} \\
            \hline
		  \specialrule{0em}{1pt}{0pt}
            GraftNet \cite{liu2022graftnet}& 30.88 & 65.91 & 0.917 \\
            \textbf{GraftNet+D3Stereo (Ours)}& \textbf{31.71} & \textbf{55.50} & \textbf{0.937} \\
            \hline
		  \specialrule{0em}{1pt}{0pt}
            HVTStereo \cite{chang2023domain}& 31.43 & 59.10 & 0.933 \\
            \textbf{HVTStereo+D3Stereo (Ours)}& \textbf{31.63} & \textbf{56.32} & \textbf{0.938}  \\
            \bottomrule [1.5pt]   	
	    \end{tabular}
        \label{tab.Real1_b}
    \end{subtable}
\end{table}

\begin{table}[t!]
	\fontsize{7}{10}\selectfont 
	\caption{Experimental results of SoTA DCNNs on the Middlebury dataset. The best results are shown in bold type.}
	\centering
	\label{tab.middlebury_1}
    {
    \setlength{\tabcolsep}{1.5mm}{
	\begin{tabular}
		{lccc}
		\toprule[1.5pt]
		\multirow{2}{*}{Method} & \multicolumn{2}{c}{{PEP (\%) $\downarrow$}} & {\multirow{2}{*}{EPE (pixel) $\downarrow$}}      \\
		\specialrule{0em}{1pt}{1pt}
		\cline{2-3}
		\specialrule{0em}{1pt}{1pt}
		&\makebox[0.065\textwidth][c]{$\delta$=1}&\makebox[0.065\textwidth][c]{$\delta$=2}\\
		\midrule[1pt]
            PSMNet \cite{PSMNet}&53.5&25.7& 5.33  \\
            \textbf{PSMNet+D3Stereo (Ours)}& \textbf{34.1} & \textbf{23.6} & \textbf{4.01}  \\
            \hline
		      \specialrule{0em}{1pt}{0pt}
            AANet \cite{AA}& 38.9 & 27.2 & 6.62  \\
            \textbf{AANet+D3Stereo (Ours)}& \textbf{31.8} & \textbf{21.2} & \textbf{4.39}  \\
            \hline
		      \specialrule{0em}{1pt}{0pt}
            BGNet \cite{BGNet}& 27.1 & 19.8 & 4.28  \\
            \textbf{BGNet+D3Stereo (Ours)}& \textbf{24.9} & \textbf{17.1} & \textbf{2.97}  \\
            \hline
		      \specialrule{0em}{1pt}{0pt}
            LacGwc \cite{LacGwc}& 27.4 & \textbf{17.1} & 4.35  \\
            \textbf{LacGwc+D3Stereo (Ours)}& \textbf{25.9} & 18.2 & \textbf{4.07}  \\
            \hline
		      \specialrule{0em}{1pt}{0pt}
            GMStereo \cite{Unimatch}& 29.7 & 19.3 & 3.39  \\
            \textbf{GMStereo+D3Stereo (Ours)}& \textbf{27.3} & \textbf{18.8} & \textbf{3.27}  \\
            \hline
		      \specialrule{0em}{1pt}{0pt}
            CreStereo \cite{Cre}& 31.2 & 22.1 & 3.77  \\
            \textbf{CreStereo+D3Stereo (Ours)}& \textbf{28.6} & \textbf{19.5} & \textbf{3.33}  \\
            \hline
            \specialrule{0em}{1pt}{0pt}
            GraftNet \cite{liu2022graftnet}& \textbf{22.7} & \textbf{12.3} & \textbf{2.67} \\
            \textbf{GraftNet+D3Stereo (Ours)}& 23.1 & 15.6 & 3.01  \\
            \hline
		\specialrule{0em}{1pt}{0pt}
            HVTStereo \cite{chang2023domain}& \textbf{21.1} & \textbf{11.8} & \textbf{2.16} \\
            \textbf{HVTStereo+D3Stereo (Ours)}& 30.5 & 21.4 & 3.35 \\
    	\bottomrule [1.5pt]   	
	\end{tabular}}}
\end{table}

The quantitative and qualitative experimental results on our created UDTIRI-Stereo dataset are presented in Table \ref{tab.SYN1} and Fig. \ref{fig.vis_syn1}, respectively. It is noticeable that when applying our proposed D3Stereo strategy to the existing stereo matching algorithms, EPE and PEP decrease by up to 63.10\% and 83.26\%, respectively. Although CreStereo demonstrates comparable performance in PEP with $\delta=1$ when adapted to the UDTIRI-Stereo dataset using D3Stereo strategy, it shows dramatic improvement in PEP with $\delta=0.5$ and EPE. Moreover, PT-D3Stereo diffuses decisive disparities across the entire image in a multi-directional fashion. This results in significantly improved disparity estimation results and a more uniform distribution of errors compared to other explicit programming-based stereo matching algorithms.

Secondly, the quantitative and qualitative experimental results on the Stereo-Road dataset are presented in Table \ref{tab.Real1} and Fig. \ref{fig.vis_real1}, respectively. It is observed that the D3Stereo strategy can improve all algorithms across different evaluation metrics, with increases ranging from 0.41\% to 10.63\% in PSNR and 0.97\% to 13.15\% in SSIM, as well as a decrease ranging from 4.19\% to 54.98\% in MSE. Additionally, we observe that by using the D3Stereo strategy, disparity estimation near or on discontinuities can be significantly improved, as highlighted with green dashed boxes in Fig. \ref{fig.vis_real1}.

It is noteworthy that GraftNet and HVTStereo achieve similar stereo matching accuracy on road scenes compared to stereo matching networks without additional domain generalization designs. Additionally, applying D3Stereo to both GraftNet and HVT-Stereo yields improved stereo matching accuracy in all metrics on both the UDTIRI-Stereo and Stereo-Road datasets. These results indicate a considerable domain gap between road scenes and other common indoor/outdoor scenes, and further demonstrate the superiority of D3Stereo in solving the road surface 3D reconstruction task. Additionally, we notice from the above experimental results that explicit programming-based algorithms developed specifically for road surface 3D reconstruction yield comparable performance to the DCNNs pre-trained on the KITTI Stereo 2015 dataset. Specifically, PT-D3Stereo outperforms most DCNNs without applying D3Stereo strategy in the majority of cases. These results suggest that the generalizability of SoTA DCNNs is still not sufficiently satisfactory for road disparity estimation. When applying our proposed D3Stereo strategy to DCNNs, a SoTA performance is achieved.

\begin{table*}[t!]
    \fontsize{7}{10}\selectfont
    \caption{Experimental results of four backbone DCNNs pre-trained on the ImageNet \cite{russakovsky2015imagenet} database. H and W denote the height and width of the input image, respectively.}
    \setlength{\tabcolsep}{1mm}
    \label{tab.backbone}
    \centering
    \begin{tabular}{llcccc}
        \toprule[1.5pt]
        \multirow{2}{*}{Backbones}  & \multirow{2}{*}{Sizes of the selected feature maps} & \multicolumn{2}{c}{UDTIRI-Stereo dataset} & \multicolumn{2}{c}{Middlebury dataset} \\
        \cmidrule(r){3-4}
        \cmidrule(r){5-6}
        \specialrule{0em}{0pt}{-3pt}
        &&EPE (pixel)$\downarrow$&Runtime (s)$\downarrow$&EPE (pixel)$\downarrow$&Runtime (s)$\downarrow$ \\
        \midrule[1pt]
        \multirow{3}{*}{VGG \cite{simonyan2014very}}& [($\frac{\text{H}}{2}$,$\frac{\text{W}}{2}$,128),\enspace($\frac{\text{H}}{4}$,$\frac{\text{W}}{4}$,256)] & 0.261 & {0.92} & 3.95 & {1.50} \\
        & [($\frac{\text{H}}{2}$,$\frac{\text{W}}{2}$,128),\enspace($\frac{\text{H}}{4}$,$\frac{\text{W}}{4}$,256),\enspace($\frac{\text{H}}{8}$,$\frac{\text{W}}{8}$,512)] & \textbf{0.241} & 0.96 & \textbf{3.75} & 1.52 \\
        & [($\frac{\text{H}}{2}$,$\frac{\text{W}}{2}$,128),\enspace($\frac{\text{H}}{4}$,$\frac{\text{W}}{4}$,256),\enspace($\frac{\text{H}}{8}$,$\frac{\text{W}}{8}$,512),\enspace($\frac{\text{H}}{16}$,$\frac{\text{W}}{16}$,512)] & 0.242 & \textbf{0.91} & 3.77 & \textbf{1.37} \\
        \specialrule{0em}{1pt}{1pt}
        \hline
        \specialrule{0em}{1pt}{0pt}
        \multirow{3}{*}{ResNet \cite{he2016deep}}& [($\frac{\text{H}}{2}$,$\frac{\text{W}}{2}$,64),\enspace($\frac{\text{H}}{4}$,$\frac{\text{W}}{4}$,64)] & 0.314 & 0.84 & 4.02 & 1.22 \\
        & [($\frac{\text{H}}{2}$,$\frac{\text{W}}{2}$,64),\enspace($\frac{\text{H}}{4}$,$\frac{\text{W}}{4}$,64),\enspace($\frac{\text{H}}{8}$,$\frac{\text{W}}{8}$,128)] & \textbf{0.309} & \textbf{0.81} & \textbf{3.88} & \textbf{1.06} \\
        & [($\frac{\text{H}}{2}$,$\frac{\text{W}}{2}$,64),\enspace($\frac{\text{H}}{4}$,$\frac{\text{W}}{4}$,64),\enspace($\frac{\text{H}}{8}$,$\frac{\text{W}}{8}$,128),\enspace($\frac{\text{H}}{16}$,$\frac{\text{W}}{16}$,256)] & 0.309 & 0.82 & 4.07 & 1.08 \\
        \specialrule{0em}{1pt}{1pt}
        \hline
        \specialrule{0em}{1pt}{0pt}
        \multirow{3}{*}{MobileNetV3 \cite{howard2019searching}}& [($\frac{\text{H}}{2}$,$\frac{\text{W}}{2}$,16),\enspace($\frac{\text{H}}{4}$,$\frac{\text{W}}{4}$,24)] & 0.672 & 0.87 & {4.67} & {1.17} \\
        & [($\frac{\text{H}}{2}$,$\frac{\text{W}}{2}$,16),\enspace($\frac{\text{H}}{4}$,$\frac{\text{W}}{4}$,24),\enspace($\frac{\text{H}}{8}$,$\frac{\text{W}}{8}$,40)] & {0.573} & {0.84} & 4.54 & 1.12 \\
        & [($\frac{\text{H}}{2}$,$\frac{\text{W}}{2}$,16),\enspace($\frac{\text{H}}{4}$,$\frac{\text{W}}{4}$,24),\enspace($\frac{\text{H}}{8}$,$\frac{\text{W}}{8}$,40),\enspace($\frac{\text{H}}{16}$,$\frac{\text{W}}{16}$,112)] & \textbf{0.562} & \textbf{0.81} & \textbf{4.48} & \textbf{1.08} \\
        \specialrule{0em}{1pt}{1pt}
        \hline
        \specialrule{0em}{1pt}{0pt}
        \multirow{3}{*}{MobileNetV3-S \cite{howard2019searching}}& [($\frac{\text{H}}{2}$,$\frac{\text{W}}{2}$,16),\enspace($\frac{\text{H}}{4}$,$\frac{\text{W}}{4}$,16)] & 1.412 & 0.89 & 7.23 & 1.05 \\
        & [($\frac{\text{H}}{2}$,$\frac{\text{W}}{2}$,16),\enspace($\frac{\text{H}}{4}$,$\frac{\text{W}}{4}$,16),\enspace($\frac{\text{H}}{8}$,$\frac{\text{W}}{8}$,24)] & \textbf{1.357} & 0.82 & 6.44 & \textbf{0.96} \\   
        & [($\frac{\text{H}}{2}$,$\frac{\text{W}}{2}$,16),\enspace($\frac{\text{H}}{4}$,$\frac{\text{W}}{4}$,16),\enspace($\frac{\text{H}}{8}$,$\frac{\text{W}}{8}$,24),\enspace($\frac{\text{H}}{16}$,$\frac{\text{W}}{16}$,48)] & 1.366 & \textbf{0.81} & \textbf{6.29} & {1.01} \\    
        \bottomrule [1.5pt]   	
    \end{tabular} 
\end{table*}

\subsection{Generalizability Evaluation for General Stereo Matching}

We further evaluate the generalizability of D3Stereo strategy for general stereo matching using the Middlebury dataset. The quantitative and qualitative experimental results are presented in Table \ref{tab.middlebury_1} and Fig. 1 in the supplementary material, respectively. The quantitative results suggest that domain generalization-aimed networks, GraftNet and HVTStereo, achieve higher stereo matching accuracy on the Middlebury dataset. However, for stereo matching networks without any domain generalization functionality incorporated, applying D3Stereo strategy to these DCNNs results in significantly improved stereo matching accuracy. Specifically, the EPE decreases by 3.54-33.69\%, while the PEP with $\delta=1$ and $\delta=2$ decreases by 2.59-36.26\% except for LacGwc, which achieves slightly lower PEP with $\delta=2$ compared with LacGwc without using D3Stereo strategy. The qualitative results indicate that these pre-trained DCNNs perform poorly in ambiguous regions where color intensities are similar, without any further model fine-tuning. In contrast, their performance improves with the use of D3Stereo strategy. 

However, our proposed D3Stereo strategy suffers from the edge-fattening issue \cite{AA,chen2023self} and exhibits limited stereo matching accuracy near/on extensive disparity discontinuities. Specifically, large disparities from foreground objects are diffused to background objects in the intra-scale decisive disparity diffusion process. This phenomenon primarily occurs at the overlaps between different objects. Therefore, we are motivated to explicitly leverage semantic segmentation results to constrain the intra-scale disparity propagation process within each spatially continuous region in our future work, thereby exploring broader applications of our proposed D3Stereo for general stereo matching.

\subsection{{Performance Evaluation on Backbone DCNNs}}
\label{sec.DCNN}

Additional experiments are conducted on the UDTIRI-Stereo and Middlebury datasets with backbone DCNNs pre-trained on the ImageNet database. The quantitative and qualitative experimental results are presented in Table \ref{tab.backbone} and Fig. 2 in the supplementary material, respectively. Surprisingly, these backbone DCNNs demonstrate the capability to perform accurate dense correspondence matching, when utilizing our proposed D3Stereo strategy. Specifically, when employing the D3Stereo strategy, VGG \cite{simonyan2014very} and ResNet \cite{he2016deep} achieve competitive results in comparison to stereo matching networks, and outperform explicit programming-based algorithms on the UDTIRI-Stereo dataset. Their achieved EPE values are even comparable to those obtained by the best-performing stereo matching networks, BGNet and CreStereo, on the Middlebury dataset. These results strongly suggest that our approach is effective in extending the capabilities of backbone DCNNs for solving the dense correspondence matching problem. Moreover, as shown in Table \ref{tab.backbone}, MobileNetV3 \cite{howard2019searching} has fewer feature channels compared to VGG and ResNet, resulting in less satisfactory stereo matching accuracy. We further validate this viewpoint with additional experiments using MobileNetV3-S \cite{howard2019searching}, a lighter version of MobileNetV3. As expected, MobileNetV3-S underperforms MobileNetV3 on both UDTIRI-Stereo and Middlebury datasets. This reinforces our belief that the number of feature channels is a crucial factor that significantly impacts the accuracy of stereo matching. Finally, it is noteworthy that these backbone DCNNs consistently achieve improved efficiency and accuracy when employing three or four feature layers instead of two feature layers. 

These results suggest that deep features obtained at 1/8 and 1/16 of the full image resolution exhibit similar performance in aggregating global information and eliminating stereo matching ambiguities. Consequently, these comparable intermediate results enable the subsequent decisive disparity diffusion and hierarchical refinement processes in D3Stereo to ultimately produce dense disparity maps with similar accuracy. Additionally, performing nearest neighbor search at 1/16 of the full image resolution significantly reduces the computational complexity compared to 1/8 of the full image resolution, which offsets the additional computational demands of decisive disparity diffusion and hierarchical refinement processes at 1/16 of the full image resolution, thereby maintaining the overall efficiency of D3Stereo.

\section{Conclusion}
\label{sec.conclusion}

This article introduces D3Stereo, a novel decisive disparity diffusion strategy. Our technical contributions include (1) a recursive bilateral filtering algorithm for efficient and adaptive cost aggregation, (2) an intra-scale disparity diffusion algorithm for sparse disparity map completion, and (3) an inter-scale disparity inheritance algorithm for fine-grained disparity estimation at higher resolution. Additionally, we also developed a new dataset to address the need for comprehensive performance quantification of stereo matching-based road surface 3D reconstruction algorithms. Through extensive experiments, we demonstrate the compatibility of D3Stereo with both explicit programming-based algorithms and pre-trained deep learning models for road disparity estimation. Furthermore, D3Stereo is also capable of adapting pre-trained networks to address stereo matching task in general scenes, as validated on the Middlebury dataset. In the future, we aim to further improve the accuracy of estimated disparities near/on disparity discontinuities and explore broader applications of D3Stereo for general stereo matching. This includes restraining the disparity diffusion process with each instance and investigating the integration of D3Stereo for self-supervised stereo matching.

\normalem
\bibliographystyle{IEEEtran}
\bibliography{refs}

\begin{thebibliography}{10}
\providecommand{\url}[1]{#1}
\csname url@samestyle\endcsname
\providecommand{\newblock}{\relax}
\providecommand{\bibinfo}[2]{#2}
\providecommand{\BIBentrySTDinterwordspacing}{\spaceskip=0pt\relax}
\providecommand{\BIBentryALTinterwordstretchfactor}{4}
\providecommand{\BIBentryALTinterwordspacing}{\spaceskip=\fontdimen2\font plus
\BIBentryALTinterwordstretchfactor\fontdimen3\font minus
  \fontdimen4\font\relax}
\providecommand{\BIBforeignlanguage}[2]{{%
\expandafter\ifx\csname l@#1\endcsname\relax
\typeout{** WARNING: IEEEtran.bst: No hyphenation pattern has been}%
\typeout{** loaded for the language `#1'. Using the pattern for}%
\typeout{** the default language instead.}%
\else
\language=\csname l@#1\endcsname
\fi
#2}}
\providecommand{\BIBdecl}{\relax}
\BIBdecl

\bibitem{ma2022computer}
N.~M. et~al., ``Computer vision for road imaging and pothole detection: a
  state-of-the-art review of systems and algorithms,'' \emph{Transportation
  Safety and Environment}, vol.~4, no.~4, p. tdac026, 2022.

\bibitem{liang2022automatic}
X.~Liang \emph{et~al.}, ``Automatic classification of pavement distress using
  {3D} ground-penetrating radar and deep convolutional neural network,''
  \emph{IEEE Transactions on Intelligent Transportation Systems}, vol.~23,
  no.~11, pp. 22\,269--22\,277, 2022.

\bibitem{fan2019road}
R.~Fan and M.~Liu, ``Road damage detection based on unsupervised disparity map
  segmentation,'' \emph{IEEE Transactions on Intelligent Transportation
  Systems}, vol.~21, no.~11, pp. 4906--4911, 2020.

\bibitem{haq2019stereo}
M.~U.~U. Haq \emph{et~al.}, ``Stereo-based {3D} reconstruction of potholes by a
  hybrid, dense matching scheme,'' \emph{IEEE Sensors Journal}, vol.~19,
  no.~10, pp. 3807--3817, 2019.

\bibitem{ahmed2021pothole}
A.~Ahmed \emph{et~al.}, ``Pothole {3D} reconstruction with a novel imaging
  system and structure from motion techniques,'' \emph{IEEE Transactions on
  Intelligent Transportation Systems}, vol.~23, no.~5, pp. 4685--4694, 2021.

\bibitem{fan2019pothole}
R.~Fan \emph{et~al.}, ``Pothole detection based on disparity transformation and
  road surface modeling,'' \emph{IEEE Transactions on Image Processing},
  vol.~29, pp. 897--908, 2019.

\bibitem{fan2018road}
R.~Fan \emph{et~al.}, ``Road surface {3D} reconstruction based on dense
  subpixel disparity map estimation,'' \emph{IEEE Transactions on Image
  Processing}, vol.~27, no.~6, pp. 3025--3035, 2018.

\bibitem{fan2021graph}
R.~Fan \emph{et~al.}, ``Graph attention layer evolves semantic segmentation for
  road pothole detection: A benchmark and algorithms,'' \emph{IEEE Transactions
  on Image Processing}, vol.~30, pp. 8144--8154, 2021.

\bibitem{liu2024playing}
C.-W. Liu \emph{et~al.}, ``Playing to vision foundation model's strengths in
  stereo matching,'' \emph{IEEE Transactions on Intelligent Vehicles}, 2024,
  {DOI}:10.1109/TIV.2024.3467287.

\bibitem{PSMNet}
J.-R. Chang and Y.-S. Chen, ``Pyramid stereo matching network,'' in
  \emph{Proceedings of the IEEE Conference on Computer Vision and Pattern
  Recognition (CVPR)}, 2018, pp. 5410--5418.

\bibitem{RAFT}
L.~Lipson \emph{et~al.}, ``{RAFT-Stereo}: Multilevel recurrent field transforms
  for stereo matching,'' in \emph{2021 International Conference on 3D Vision
  (3DV)}.\hskip 1em plus 0.5em minus 0.4em\relax IEEE, 2021, pp. 218--227.

\bibitem{liu2023stereo}
C.-W. Liu \emph{et~al.}, ``Stereo matching: fundamentals, state-of-the-art, and
  existing challenges,'' in \emph{Autonomous Driving Perception: Fundamentals
  and Applications}.\hskip 1em plus 0.5em minus 0.4em\relax Springer, 2023, pp.
  63--100.

\bibitem{FBS}
R.~Fan \emph{et~al.}, ``Real-time dense stereo embedded in a {UAV} for road
  inspection,'' in \emph{Proceedings of the IEEE/CVF Conference on Computer
  Vision and Pattern Recognition Workshops (CVPRW)}, 2019, pp. 535--543.

\bibitem{fan2021rethinking}
R.~Fan \emph{et~al.}, ``Rethinking road surface {3-D} reconstruction and
  pothole detection: From perspective transformation to disparity map
  segmentation,'' \emph{IEEE Transactions on Cybernetics}, vol.~52, no.~7, pp.
  5799--5808, 2022.

\bibitem{roy1999stereo}
S.~Roy, ``Stereo without epipolar lines: A maximum-flow formulation,''
  \emph{International Journal of Computer Vision}, vol.~34, no. 2-3, pp.
  147--161, 1999.

\bibitem{miksik2015incremental}
O.~Miksik \emph{et~al.}, ``Incremental dense multi-modal {3D} scene
  reconstruction,'' in \emph{IEEE/RSJ International Conference on Intelligent
  Robots and Systems (IROS)}.\hskip 1em plus 0.5em minus 0.4em\relax IEEE,
  2015, pp. 908--915.

\bibitem{pillai2016high}
S.~Pillai \emph{et~al.}, ``High-performance and tunable stereo
  reconstruction,'' in \emph{IEEE International Conference on Robotics and
  Automation (ICRA)}.\hskip 1em plus 0.5em minus 0.4em\relax IEEE, 2016, pp.
  3188--3195.

\bibitem{efe2021dfm}
U.~Efe \emph{et~al.}, ``{DFM}: {A} performance baseline for deep feature
  matching,'' in \emph{Proceedings of the IEEE/CVF Conference on Computer
  Vision and Rattern Recognition Workshop (CVPRW)}, 2021, pp. 4284--4293.

\bibitem{zhou2023e3cm}
C.~Zhou \emph{et~al.}, ``{E3CM}: Epipolar-constrained cascade correspondence
  matching,'' \emph{Neurocomputing}, vol. 559, p. 126788, 2023.

\bibitem{russakovsky2015imagenet}
O.~Russakovsky \emph{et~al.}, ``Image{N}et large scale visual recognition
  challenge,'' \emph{International Journal of Computer Vision}, vol. 115, pp.
  211--252, 2015.

\bibitem{dosovitskiy2017carla}
A.~Dosovitskiy \emph{et~al.}, ``{CARLA}: {A}n open urban driving simulator,''
  in \emph{Conference on Robot Learning (CoRL)}.\hskip 1em plus 0.5em minus
  0.4em\relax PMLR, 2017, pp. 1--16.

\bibitem{cabon2020virtual}
\BIBentryALTinterwordspacing
Y.~Cabon \emph{et~al.}, ``Virtual {KITTI} 2,'' \emph{Computing Research
  Repository {(CoRR)}}, vol. abs/2001.10773, 2020. [Online]. Available:
  \url{https://arxiv.org/abs/2001.10773}
\BIBentrySTDinterwordspacing

\bibitem{Cre}
J.~Li \emph{et~al.}, ``Practical stereo matching via cascaded recurrent network
  with adaptive correlation,'' in \emph{Proceedings of the IEEE/CVF Conference
  on Computer Vision and Pattern Recognition (CVPR)}, 2022, pp.
  16\,263--16\,272.

\bibitem{zhang2014efficient}
Z.~Zhang \emph{et~al.}, ``An efficient algorithm for pothole detection using
  stereo vision,'' in \emph{2014 IEEE International Conference on Acoustics,
  Speech and Signal Processing (ICASSP)}.\hskip 1em plus 0.5em minus
  0.4em\relax IEEE, 2014, pp. 564--568.

\bibitem{rao2023masked}
Z.~Rao \emph{et~al.}, ``Masked representation learning for domain generalized
  stereo matching,'' in \emph{Proceedings of the IEEE/CVF Conference on
  Computer Vision and Pattern Recognition (CVPR)}, 2023, pp. 5435--5444.

\bibitem{chang2023domain}
T.~Chang \emph{et~al.}, ``Domain generalized stereo matching via hierarchical
  visual transformation,'' in \emph{Proceedings of the IEEE/CVF Conference on
  Computer Vision and Pattern Recognition (CVPR)}, 2023, pp. 9559--9568.

\bibitem{zhang2022revisiting}
J.~Zhang \emph{et~al.}, ``Revisiting domain generalized stereo matching
  networks from a feature consistency perspective,'' in \emph{Proceedings of
  the IEEE/CVF Conference on Computer Vision and Pattern Recognition (CVPR)},
  2022, pp. 13\,001--13\,011.

\bibitem{liu2022graftnet}
B.~Liu \emph{et~al.}, ``{GraftNet}: Towards domain generalized stereo matching
  with a broad-spectrum and task-oriented feature,'' in \emph{Proceedings of
  the IEEE/CVF Conference on Computer Vision and Pattern Recognition (CVPR)},
  2022, pp. 13\,012--13\,021.

\bibitem{shen2021cfnet}
Z.~Shen \emph{et~al.}, ``{CFNet}: Cascade and fused cost volume for robust
  stereo matching,'' in \emph{Proceedings of the IEEE/CVF Conference on
  Computer Vision and Pattern Recognition (CVPR)}, 2021, pp. 13\,906--13\,915.

\bibitem{lowe2004distinctive}
D.~G. Lowe, ``Distinctive image features from scale-invariant keypoints,''
  \emph{International Journal of Computer Vision}, vol.~60, pp. 91--110, 2004.

\bibitem{bay2006surf}
H.~Bay \emph{et~al.}, ``{SURF}: Speeded up robust features,'' in \emph{European
  Conference on Computer Vision (ECCV)}.\hskip 1em plus 0.5em minus 0.4em\relax
  Springer, 2006, pp. 404--417.

\bibitem{leutenegger2011brisk}
S.~Leutenegger \emph{et~al.}, ``{BRISK}: {B}inary robust invariant scalable
  keypoints,'' in \emph{International Conference on Computer Vision
  (ICCV)}.\hskip 1em plus 0.5em minus 0.4em\relax IEEE, 2011, pp. 2548--2555.

\bibitem{barroso2022key}
A.~Barroso-Laguna and K.~Mikolajczyk, ``{K}ey. {N}et: Keypoint detection by
  handcrafted and learned {CNN} filters revisited,'' \emph{IEEE Transactions on
  Pattern Analysis and Machine Intelligence}, vol.~45, no.~1, pp. 698--711,
  2022.

\bibitem{detone2018superpoint}
D.~DeTone \emph{et~al.}, ``{SuperPoint}: Self-supervised interest point
  detection and description,'' in \emph{Proceedings of the IEEE Conference on
  Computer Vision and Pattern Recognition Workshops (CVPRW)}, 2018, pp.
  224--236.

\bibitem{dusmanu2019d2}
M.~Dusmanu \emph{et~al.}, ``{D2-Net}: A trainable {CNN} for joint description
  and detection of local features,'' in \emph{Proceedings of the IEEE/CVF
  Conference on Computer Vision and Pattern Recognition (CVPR)}, 2019, pp.
  8092--8101.

\bibitem{revaud2019r2d2}
J.~Revaud \emph{et~al.}, ``{R2D2}: Reliable and repeatable detector and
  descriptor,'' \emph{Advances in Neural Information Processing Systems
  (NeurIPS)}, vol.~32, pp. 12\,405--12\,415, 2019.

\bibitem{sarlin2020superglue}
P.-E. Sarlin \emph{et~al.}, ``Super{G}lue: Learning feature matching with graph
  neural networks,'' in \emph{Proceedings of the IEEE/CVF Conference on
  Computer Vision and Pattern Recognition (CVPR)}, 2020, pp. 4938--4947.

\bibitem{rocco2020ncnet}
I.~Rocco \emph{et~al.}, ``{NCNet}: Neighbourhood consensus networks for
  estimating image correspondences,'' \emph{IEEE Transactions on Pattern
  Analysis and Machine Intelligence}, vol.~44, no.~2, pp. 1020--1034, 2020.

\bibitem{rocco2020efficient}
I.~Rocco \emph{et~al.}, ``Efficient neighborhood consensus networks via
  submanifold sparse convolutions,'' in \emph{European Conference on Computer
  Vision (ECCV)}.\hskip 1em plus 0.5em minus 0.4em\relax Springer, 2020, pp.
  605--621.

\bibitem{sun2021loftr}
J.~Sun \emph{et~al.}, ``{LoFTR}: Detector-free local feature matching with
  {Transformers},'' in \emph{Proceedings of the IEEE/CVF Conference on Computer
  Vision and Pattern Recognition (CVPR)}, 2021, pp. 8922--8931.

\bibitem{tappen2003comparison}
M.~F. Tappen and W.~T. Freeman, ``Comparison of graph cuts with belief
  propagation for stereo, using identical {MRF} parameters,'' in
  \emph{Proceedings of the IEEE International Conference on Computer Vision
  (ICCV)}.\hskip 1em plus 0.5em minus 0.4em\relax IEEE, 2003, pp. 900--906.

\bibitem{wang2021pvstereo}
H.~Wang \emph{et~al.}, ``{PVStereo}: {P}yramid voting module for end-to-end
  self-supervised stereo matching,'' \emph{IEEE Robotics and Automation
  Letters}, vol.~6, no.~3, pp. 4353--4360, 2021.

\bibitem{zhang2017cross}
K.~Zhang \emph{et~al.}, ``Cross-scale cost aggregation for stereo matching,''
  \emph{IEEE Transactions on Circuits and Systems for Video Technology},
  vol.~27, no.~5, pp. 965--976, 2017.

\bibitem{mozerov2015accurate}
M.~G. Mozerov and J.~Van De~Weijer, ``Accurate stereo matching by two-step
  energy minimization,'' \emph{IEEE Transactions on Image Processing}, vol.~24,
  no.~3, pp. 1153--1163, 2015.

\bibitem{simonyan2014very}
K.~Simonyan and A.~Zisserman, ``Very deep convolutional networks for
  large-scale image recognition,'' in \emph{International Conference on
  Learning Representations (ICLR)}, 2015.

\bibitem{radosavovic2020designing}
I.~Radosavovic \emph{et~al.}, ``Designing network design spaces,'' in
  \emph{Proceedings of the IEEE/CVF Conference on Computer Vision and Pattern
  Recognition (CVPR)}, 2020, pp. 10\,428--10\,436.

\bibitem{luo2016understanding}
W.~Luo \emph{et~al.}, ``Understanding the effective receptive field in deep
  convolutional neural networks,'' \emph{Advances in Neural Information
  Processing Systems (NeurIPS)}, vol.~29, pp. 4898--4906, 2016.

\bibitem{poggi2021confidence}
M.~Poggi \emph{et~al.}, ``On the confidence of stereo matching in a
  deep-learning era: a quantitative evaluation,'' \emph{IEEE Transactions on
  Pattern Analysis and Machine Intelligence}, vol.~44, no.~9, pp. 5293--5313,
  2021.

\bibitem{xu2023iterative}
G.~Xu \emph{et~al.}, ``Iterative geometry encoding volume for stereo
  matching,'' in \emph{Proceedings of the IEEE/CVF Conference on Computer
  Vision and Pattern Recognition (CVPR)}, 2023, pp. 21\,919--21\,928.

\bibitem{zhou2021patch2pix}
Q.~Zhou \emph{et~al.}, ``{P}atch2{P}x: {E}pipolar-guided pixel-level
  correspondences,'' in \emph{Proceedings of the IEEE/CVF Conference on
  Computer Vision and Pattern Recognition (CVPR)}, 2021, pp. 4669--4678.

\bibitem{scharstein2014high}
D.~Scharstein \emph{et~al.}, ``High-resolution stereo datasets with
  subpixel-accurate ground truth,'' in \emph{German Conference on Pattern
  Recognition (GCPR)}.\hskip 1em plus 0.5em minus 0.4em\relax Springer, 2014,
  pp. 31--42.

\bibitem{AA}
H.~Xu and J.~Zhang, ``{AANet}: Adaptive aggregation network for efficient
  stereo matching,'' in \emph{Proceedings of the IEEE/CVF Conference on
  Computer Vision and Pattern Recognition (CVPR)}, 2020, pp. 1959--1968.

\bibitem{BGNet}
B.~Xu \emph{et~al.}, ``Bilateral grid learning for stereo matching networks,''
  in \emph{Proceedings of the IEEE/CVF Conference on Computer Vision and
  Pattern Recognition (CVPR)}, 2021, pp. 12\,497--12\,506.

\bibitem{LacGwc}
B.~Liu \emph{et~al.}, ``Local similarity pattern and cost self-reassembling for
  deep stereo matching networks,'' in \emph{Proceedings of the AAAI Conference
  on Artificial Intelligence (AAAI)}, 2022, pp. 1647--1655.

\bibitem{Unimatch}
H.~Xu \emph{et~al.}, ``Unifying flow, stereo and depth estimation,'' \emph{IEEE
  Transactions on Pattern Analysis and Machine Intelligence}, vol.~45, no.~11,
  pp. 13\,941--13\,958, 2023.

\bibitem{menze2015object}
M.~Menze and A.~Geiger, ``Object scene flow for autonomous vehicles,'' in
  \emph{Proceedings of the IEEE Conference on Computer Vision and Pattern
  Recognition (CVPR)}, 2015, pp. 3061--3070.

\bibitem{mayer2016large}
N.~Mayer \emph{et~al.}, ``A large dataset to train convolutional networks for
  disparity, optical flow, and scene flow estimation,'' in \emph{Proceedings of
  the IEEE Conference on Computer Vision and Pattern Recognition (CVPR)}, 2016,
  pp. 4040--4048.

\bibitem{he2016deep}
K.~He \emph{et~al.}, ``Deep residual learning for image recognition,'' in
  \emph{Proceedings of the IEEE Conference on Computer Vision and Pattern
  Recognition (CVPR)}, 2016, pp. 770--778.

\bibitem{howard2019searching}
A.~Howard \emph{et~al.}, ``Searching for {MobileNetV3},'' in \emph{Proceedings
  of the IEEE/CVF International Conference on Computer Vision (ICCV)}, 2019,
  pp. 1314--1324.

\bibitem{chen2023self}
X.~Chen \emph{et~al.}, ``Self-supervised monocular depth estimation: Solving
  the edge-fattening problem,'' in \emph{Proceedings of the IEEE/CVF Winter
  Conference on Applications of Computer Vision (WACV)}, 2023, pp. 5776--5786.

\end{thebibliography}

\end{document}